\documentclass[11pt]{article}
\usepackage[margin=1in]{geometry}
\usepackage[T1]{fontenc}
\usepackage{lmodern}
\usepackage{amsmath,amssymb}
\usepackage{booktabs}
\usepackage{placeins}
\usepackage{amsthm}
\usepackage{algorithm}
\usepackage{algorithmic}
\newtheorem{proposition}{Proposition}
\newcommand{\codestatement}{}
\usepackage{graphicx}
\usepackage{pgfplots}
\usepgfplotslibrary{groupplots}
\pgfplotsset{compat=1.17}
\usepackage[numbers,sort&compress]{natbib}
\usepackage[colorlinks=true,linkcolor=blue!50!black,citecolor=blue!50!black,urlcolor=blue!50!black]{hyperref}

\title{Find, Count and Fix Wrong Labels from a Trained Model,\\ without Clean Labels or Shared Data}
\author{Yuchen Li
  \and Zongqi Fan$^{*}$ \and Xintong Li$^{*}$ \and Nguyen H. Tran$^{\dagger}$ \and Ken-Tye Yong$^{\dagger}$\\[4pt]
  {\small The University of Sydney}\\[2pt]
  {\small $^{*}$Joint second authors. \quad $^{\dagger}$Corresponding authors.}}
\date{Preliminary version, November 2026}

\begin{document}
\maketitle

\begin{abstract}
How many labels in a training set are wrong, and which ones? The question arises when labels are accepted from
annotators and when a trained model is taken over with its training data, and readings from a model's outputs fail
once it has fitted its wrong labels. We read the labels from the per-example gradients of the trained model.
Normalised into a \emph{gradient allocation}, which records where an example asks the network to change, the gradient
of a wrongly labelled example resembles its true class even after the label is fitted. An example is flagged when
relabelling it as its nearest other class fits its allocation better, with no threshold, clean data, held-out set or
further training. After the model has fitted 99--100\% of its wrong labels, the audit separates wrong from right labels
with an AUC of 0.94--0.96 on the noisiest human labels of CIFAR-10N, where the loss, gradient norm and self-influence
fall to 0.73--0.76. It is the highest of the scores that read only the final model in all 22 image and 9 text settings
and with 1{,}000 classes. Its reading is a lower bound on the noise rate, and the examples it flags are almost all wrong,
while readings from predictions or feature margins fall to 0--6\%. Retraining without the flagged examples raises the
test accuracy by up to 47 points. The data owner computes the few numbers per example that the audit needs, from which
a trained decoder recovers no more than the mean image of the class. Auditing 5{,}000 images takes 14 to 27 seconds on
one GPU.
\end{abstract}

\section{Introduction}

Training labels are often wrong. Labels collected from people carry the confusions of their annotators
\citep{wei2022cifarn}, and labels collected from the web carry the errors of the pages they came from
\citep{xiao2015clothing1m}. A network trained long enough fits these labels \citep{zhang2017rethinking,arpit2017closer}.
How many labels are wrong, and which ones, then matters when a batch of labels is accepted from annotators, when a
trained model is taken over with its training set, and when examples are chosen for re-annotation. In each case the
trained model and its labelled training examples are at hand, while clean labels, a held-out set, the checkpoints of
training and its recipe often are not. When the data cannot be handed over, as with sensitive records whose labels a
buyer or a regulator wants to check, the answer has to come from statistics that the owner computes on its own
machine, as in federated learning \citep{mcmahan2017fedavg}.

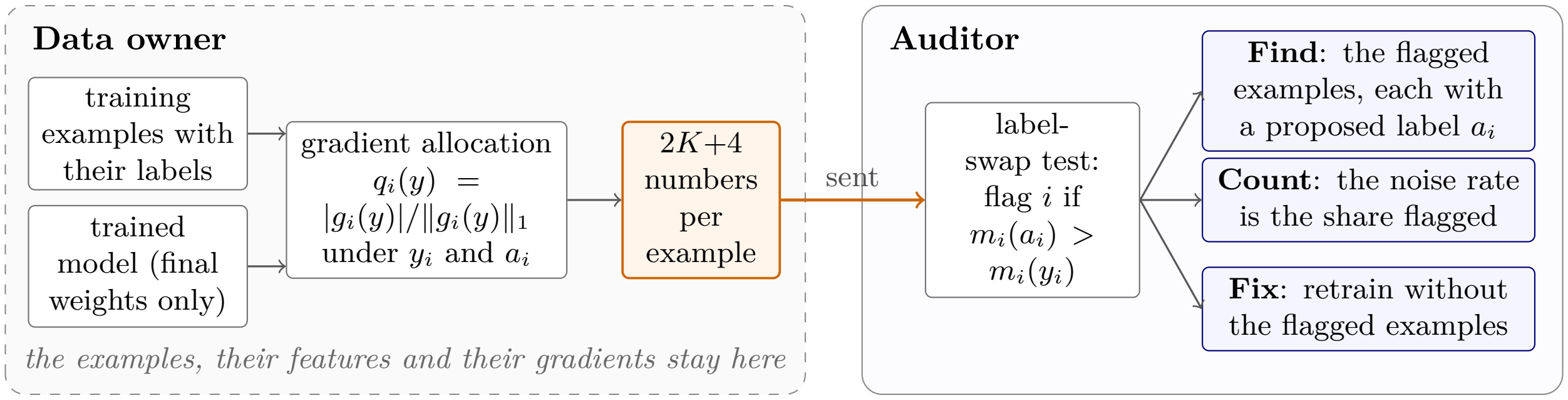
\begin{figure*}[t]
\centering
\begin{tikzpicture}[x=1cm, y=1cm, font=\footnotesize,
  box/.style={draw=black!55, rounded corners=2pt, align=center, inner sep=3pt, fill=white},
  key/.style={box, draw=orange!80!black, fill=orange!8, line width=0.7pt},
  outbox/.style={box, text width=3.3cm, fill=blue!4, draw=blue!45!black},
  arr/.style={->, line width=0.6pt, black!65}]
\draw[dashed, rounded corners=6pt, black!45, fill=black!2] (0,0) rectangle (8.45,4.1);
\node[anchor=north west, font=\small\bfseries] at (0.15,4.02) {Data owner};
\node[anchor=south, font=\footnotesize\itshape, black!60] at (4.22,0.08) {the examples, their features and their gradients stay here};
\draw[rounded corners=6pt, black!45, fill=blue!1] (9.05,0) rectangle (16.45,4.1);
\node[anchor=north west, font=\small\bfseries] at (9.2,4.02) {Auditor};
\node[box, text width=2.1cm] (data) at (1.4,2.75) {training examples with their labels};
\node[box, text width=2.1cm] (model) at (1.4,1.35) {trained model (final weights only)};
\node[box, text width=2.75cm] (alloc) at (4.45,2.05) {gradient allocation\\ $q_i(y) = |g_i(y)| / \lVert g_i(y) \rVert_1$\\ under $y_i$ and $a_i$};
\node[key, text width=1.45cm] (msg) at (7.35,2.05) {$2K{+}4$ numbers per example};
\draw[arr] (data.east) -- (alloc.west |- data.east);
\draw[arr] (model.east) -- (alloc.west |- model.east);
\draw[arr] (alloc) -- (msg);
\node[box, text width=2.05cm] (test) at (10.85,2.05) {label-swap test:\\ flag $i$ if\\ $m_i(a_i) > m_i(y_i)$};
\draw[arr, line width=0.9pt, orange!80!black] (msg) -- node[above, font=\footnotesize, black!70] {sent} (test);
\node[outbox] (find) at (14.4,3.2) {\textbf{Find}: the flagged examples, each with a proposed label $a_i$};
\node[outbox] (count) at (14.4,2.05) {\textbf{Count}: the noise rate is the share flagged};
\node[outbox] (fix) at (14.4,0.9) {\textbf{Fix}: retrain without the flagged examples};
\draw[arr] (test.east) -- (find.west);
\draw[arr] (test.east) -- (count.west);
\draw[arr] (test.east) -- (fix.west);
\end{tikzpicture}
\caption{The audit at a glance. The owner of the data runs the trained model on its own examples and computes, for
each example, the gradient allocation under its observed label $y_i$ and under the closest other class $a_i$
(Section~\ref{sec:method}), and sends $2K + 4$ numbers per example, with the norm and the size of each class sum. The
auditor applies the label-swap test to these numbers alone and obtains the three outputs. No clean labels, held-out
data, checkpoints or further training are needed, and the examples, their features and their gradients stay with
the owner.}
\label{fig:overview}
\end{figure*}

Once a model has fitted a wrong label, the usual readings lose their signal. The model gives the label a small loss and
predicts it, so scores built on the loss, the confidence or the prediction
\citep{northcutt2021confident,pleiss2020aum,swayamdipta2020cartography} see a clean training set, and the gradient norm
and the self-influence of an example \citep{paul2021el2n,pruthi2020tracin} shrink together with the loss. The readings
that keep their signal need more than the trained model: checkpoints saved during training, further trainings, or a
held-out set, whose reading also depends on how that set was collected (Appendix~\ref{app:further}).

Our starting point is an account of how fine-tuning uses the features of a pretrained network \citep{li2026whygap}. In
face recognition, all faces share an outline and eyes, groups of faces share an age or a pose, and a single face has
marks of its own. Training first fits what many examples share and later what few of them have. The per-example
gradients, normalised into a \emph{gradient allocation} that records where an example asks the network to change, show
through which features an example is fitted. A correctly labelled example is fitted through what its class shares, so
its allocation lies with the allocations of its class. A wrongly labelled example lacks the features that its labelled
class shares, so its allocation lies apart, and it returns to its true class once it is relabelled. Fitting the label
shrinks the size of the gradient but not its pattern, and the allocation normalises the size away
(Section~\ref{sec:method}). The signal thus lies inside the audited model and survives memorisation.

The \emph{label-swap audit} turns this into a decision. Each example is relabelled as the class whose mean allocation
it resembles most, and it is flagged when its allocation fits that class better than its own label. The share of
flagged examples is the audit's reading of the noise rate, and the flagged examples with their new labels are proposed
corrections. The rule compares two numbers of the same example and has no threshold. It needs only the trained model
and its labelled training examples, and it audits 5{,}000 examples in 14 to 27 seconds on one GPU, about one hundredth
of the time it took to train the model (Figure~\ref{fig:overview}).

Our contributions are the following.
\begin{itemize}
\item \textbf{A signal that survives memorisation.} After the model has fitted 99--100\% of its wrong labels, the
allocation separates wrong from right labels with an AUC of 0.94--0.96 on the noisiest human labels of CIFAR-10N, where
the loss, the gradient norm and the self-influence fall to 0.73--0.76. It is the highest of the scores that read only
the final model in all 22 noisy image settings and all nine text settings, and in every run under asymmetric noise and
with 1{,}000 classes (Sections~\ref{sec:survive}, \ref{sec:text} and~\ref{sec:cost}).
\item \textbf{A lower bound on the noise rate that can be used to accept labels.} Once the wrong labels are fitted,
readings from predictions and feature margins fall to 0--6\% at true noise rates of 8--58\% (ViT-S, CIFAR-10). The
audit reads 54--96\% of the true rate there, at least 93\% of the examples it counts carry wrong labels, and it flags at
most 1.0\% of a clean training set. The wrong labels it misses are mostly those that are easy to confuse
(Section~\ref{sec:reading}).
\item \textbf{An audit on the Pareto front of cost and accuracy.} The audit reads a model as it is handed over: no
clean labels, no held-out set, no checkpoints and no further training. The scores that match or exceed it in some
settings need checkpoints from the training run (TracIn) or four further trainings (confident learning, at about 100
times the cost of the audit) (Sections~\ref{sec:cl} and~\ref{sec:cost}).
\item \textbf{Flags that can be acted on.} Retraining without the flagged examples raises the test accuracy in all 22
noisy settings, by 11.5 points on average and up to 47 points, while removing as many random examples does not. It is
level with the cleaning of confident learning, without its further trainings (Section~\ref{sec:retrain}).
\item \textbf{An audit that does not move the data.} The decision for an example needs $2K + 4$ numbers for $K$
classes, which the owner of the data computes on its own machine. These numbers protect the data well: a decoder
trained to invert them reconstructs images no closer to the originals than the mean image of their class, and
re-identifies fewer than a quarter as many images as from the final features (Section~\ref{sec:private}).
\end{itemize}

\section{Related work}
\label{sec:related}

\paragraph{Finding mislabelled examples.} Confident learning estimates the joint distribution of observed and true
labels from out-of-sample probabilities, which takes further trainings \citep{northcutt2021confident}.
Area-under-the-margin \citep{pleiss2020aum} and dataset cartography \citep{swayamdipta2020cartography} need records of
the whole training run. Feature-space methods compare an example with its class or its neighbours \citep{kim2021fine},
and SimiFeat does so in the feature space of an external pretrained encoder \citep{zhu2022simifeat}. Gradient-based
methods select robust coresets \citep{mirzasoleiman2020crust}, score gradient norms early in training
\citep{paul2021el2n} or sum the self-influence over checkpoints \citep{pruthi2020tracin}. Our audit reads one frozen
model after memorisation and compares normalised allocations of the same example under two labels. The allocation was
introduced to explain the train--validation gap of fine-tuning \citep{li2026whygap}.

\paragraph{Learning with noisy labels.} Robust losses \citep{zhang2018gce}, small-loss selection
\citep{han2018coteaching,li2020dividemix} and early-learning regularisation \citep{liu2020elr} change how a model is
trained, and human noise is harder for them than synthetic noise of the same rate \citep{wei2022cifarn}. Our audit
measures the labels of a model that has already been trained.

\paragraph{Federated learning and gradient leakage.} Federated learning keeps the data with their owners
\citep{mcmahan2017fedavg}, and gradient-inversion attacks reconstruct the inputs from the gradients that are shared
\citep{zhu2019dlg,geiping2020inverting}. The split audit sends no gradients, and we measure what a decoder recovers from
the numbers it does send (Section~\ref{sec:private}).

\section{Method}
\label{sec:method}

The audit reads the frozen model, the audited training examples and their observed labels. It uses no other model, no
clean labels, no held-out or validation data and no test data. In our experiments the true labels only score the audit
afterwards.

\paragraph{Gradient allocation.}
Let $g_i(y)$ be the gradient of the cross-entropy loss of training example $x_i$ with label $y$, restricted to a set of
monitored parameters: the output projections of the last four blocks of the backbone (Appendix~\ref{app:protocol}). The
model is in evaluation mode, so the examples of a batch do not interact, also in a network with batch normalisation, and
one backward pass gives exact per-example gradients (Appendix~\ref{app:impl}). The allocation of the example under label
$y$ is the normalised magnitude vector
\begin{equation}
q_i(y) = \frac{|g_i(y)|}{\lVert g_i(y) \rVert_1},
\end{equation}
taken element-wise. It records where the example asks the network to change, and not how strongly.

\paragraph{Allocation margin.}
Let $\mu_c$ be the mean of $q_j(y_j)$ over the audited examples whose observed label $y_j$ is $c$. The margin of
example $i$ under label $y$ is
\begin{equation}
m_i(y) = \cos\bigl(q_i(y), \mu_y\bigr) - \max_{c \neq y} \cos\bigl(q_i(y), \mu_c\bigr),
\label{eq:margin}
\end{equation}
where example $i$ is left out of its own class mean when $y = y_i$. A negative margin $m_i(y_i)$ means that the
example's allocation under its observed label resembles another class more than its own.

\paragraph{Label-swap audit.}
For each example let $a_i = \arg\max_{c \neq y_i} \cos(q_i(y_i), \mu_c)$ be the closest other class. We recompute the
allocation with $a_i$ as the label and flag the example when
\begin{equation}
m_i(a_i) > m_i(y_i).
\label{eq:swap}
\end{equation}
The \emph{reading} of the audit is the share of flagged examples, and the flagged examples together with $a_i$ form a
list of suspected errors with proposed corrections. The rule compares two numbers for the same example and has no
threshold. It uses the observed labels only through the class means.

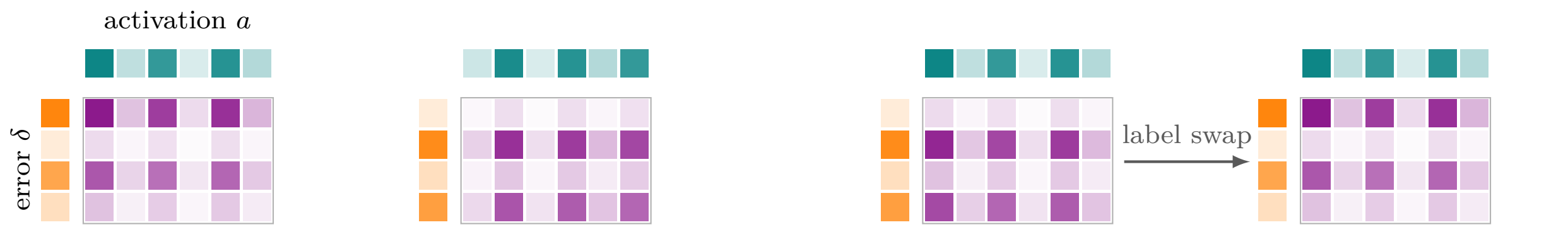
\begin{figure*}[t]
\centering
\begin{tikzpicture}[x=1.1cm, y=1.1cm]
\newcommand{\allocpanel}[6]{%
  \foreach \a [count=\c] in #4 {
    \pgfmathtruncatemacro{\v}{round(100*\a)}
    \fill[#6!\v!white] (#1+0.30*\c-0.30, #2+0.20) rectangle ++(0.27, 0.27);
  }
  \foreach \d [count=\r] in #3 {
    \pgfmathtruncatemacro{\v}{round(100*\d)}
    \fill[#5!\v!white] (#1-0.42, #2-0.30*\r+0.03) rectangle ++(0.27, 0.27);
    \foreach \a [count=\c] in #4 {
      \pgfmathtruncatemacro{\v}{round(100*\d*\a)}
      \fill[violet!\v!white] (#1+0.30*\c-0.30, #2-0.30*\r+0.03) rectangle ++(0.27, 0.27);
    }
  }
  \draw[gray!60] (#1-0.02, #2-1.19) rectangle (#1+1.79, #2+0.01);
}
\def\acat{0.95,0.25,0.80,0.15,0.85,0.30}
\def\adog{0.20,0.90,0.15,0.85,0.30,0.80}
\def\ecat{0.95,0.15,0.70,0.25}
\def\edog{0.15,0.90,0.25,0.75}
\allocpanel{0.0}{0}{\ecat}{\acat}{orange}{teal}
\allocpanel{3.6}{0}{\edog}{\adog}{orange}{teal}
\allocpanel{8.0}{0}{\edog}{\acat}{orange}{teal}
\allocpanel{11.6}{0}{\ecat}{\acat}{orange}{teal}
\node[font=\scriptsize, align=center, anchor=north] at (0.88, -1.35) {(a) mean allocation\\of the class \emph{cat}};
\node[font=\scriptsize, align=center, anchor=north] at (4.48, -1.35) {(b) mean allocation\\of the class \emph{dog}};
\node[font=\scriptsize, align=center, anchor=north] at (8.88, -1.35) {(c) a cat labelled \emph{dog}:\\cat activations, \emph{dog} errors};
\node[font=\scriptsize, align=center, anchor=north] at (12.48, -1.35) {(d) the same cat relabelled\\\emph{cat}: matches (a)};
\node[font=\scriptsize, anchor=south] at (0.88, 0.55) {activation $a$};
\node[font=\scriptsize, anchor=east, rotate=90] at (-0.62, -0.15) {error $\delta$};
\draw[-latex, thick, gray!70!black] (9.9, -0.6) -- node[font=\scriptsize, above] {label swap} (11.1, -0.6);
\end{tikzpicture}
\caption{Why the allocation survives fitting (schematic). The gradient of a monitored projection is a sum of outer
products of the error $\delta$ returned to its output and its input activation $a$; a shade is the product of the
two. Fitting a wrong label shrinks $\delta$, which the normalisation removes. What remains combines the activations of
the input with the error pattern of its label: a cat labelled \emph{dog} (c) matches neither class mean, and the same
cat relabelled \emph{cat} (d) matches the class \emph{cat} (a), which is what the label-swap test detects.}
\label{fig:mechanism}
\end{figure*}

\paragraph{Why the allocation survives fitting.}
Fitting a label shrinks the gradient of its example, and the allocation does not see this.
\begin{proposition}
\label{prop:scale}
Let $f_c$ be the logits of the model, $p = \mathrm{softmax}(f(x_i))$ and $\ell_i(y) = -\log p_y$. For any monitored
parameters $\theta$ and any label $y$ with $p_y < 1$,
\begin{equation}
\nabla_\theta \ell_i(y) = -(1 - p_y)\, \nabla_\theta \varphi_i^{y},
\label{eq:scale}
\end{equation}
where $\varphi_i^{y} = f_y(x_i) - \sum_{c \neq y} r_c f_c(x_i)$ and the weights $r_c = p_c / (1 - p_y)$ are held fixed
in the derivative. Hence, whenever $\nabla_\theta \varphi_i^{y} \neq 0$,
\begin{equation}
q_i(y) = \frac{|\nabla_\theta \varphi_i^{y}|}{\lVert \nabla_\theta \varphi_i^{y} \rVert_1}.
\end{equation}
\end{proposition}
The proof is one line: $\nabla_\theta \ell_i(y) = \sum_c (p_c - \mathbf{1}[c = y]) \nabla_\theta f_c$, and the positive
factor $1 - p_y$ cancels in the normalisation. The confidence $1 - p_y$, which fitting drives to zero, therefore enters
only the size of the gradient. This is why the loss, the gradient norm and the self-influence of a fitted wrong label
come to look like those of a right one, while the allocation stays that of the margin $\varphi_i^{y}$ between the label
and the other classes. For two classes this is the invariance shown by \citet{li2026whygap}. What the allocation
records combines the activations of the input, which carry its content, with the error pattern of its label. A cat
labelled as a dog therefore resembles neither the dogs nor the cats, and relabelling it as a cat restores both factors,
which is what the label-swap test measures (Figure~\ref{fig:mechanism}). Appendix~\ref{app:impl} gives the two factors
and shows that the second pass of the test is unaffected by fitting as well.

\paragraph{Auditing data that stays with its owner.}
The computation splits between the owner of the data and an auditor. The owner runs the frozen model on its examples,
forms the class sums $S_c$ of the allocations under the observed labels, and for each example computes the inner
products of $q_i(y_i)$ and of $q_i(a_i)$ with every $S_c$, the two squared norms $\lVert q_i(y_i) \rVert^2$ and
$\lVert q_i(a_i) \rVert^2$, and the labels $y_i$ and $a_i$: $2K + 4$ numbers per example, plus the norm and the size of
each class sum. These are all that Eqs.~\eqref{eq:margin} and~\eqref{eq:swap} need, including leaving the example out
of its own class mean, so the auditor reproduces every margin, flag and the reading exactly
(Algorithm~\ref{alg:split} in Appendix~\ref{app:algorithm}). Every label-swap result in this paper was computed from
these statistics alone. The examples, their features and their gradients stay with the owner, as the data do in
federated learning \citep{mcmahan2017fedavg}. The split audit assumes an owner that computes these statistics as
specified, as a participant in federated learning does. What it protects is the owner's data from the auditor, and
Section~\ref{sec:private} measures how well.

\section{Experiments}
\label{sec:experiments}

\subsection{Setup}
\label{sec:setup}

\paragraph{Data.} We use the human labels of CIFAR-10N \citep{wei2022cifarn} named \emph{aggregate}, \emph{random-1}
and \emph{worst} (9.0\%, 17.2\% and 40.2\% wrong) and of CIFAR-100N (40.2\%), symmetric noise at 20\% and 60\% on
CIFAR-10 and CIFAR-100, and clean CIFAR-10 and CIFAR-100. CIFAR-10H \citep{peterson2019human} gives one randomly drawn
annotator per image (5.1\% wrong). Food-101 \citep{bossard2014food} carries 40\% symmetric noise, and Food-101N
\citep{lee2018cleannet} its web labels (18.4\% wrong). Section~\ref{sec:cost} adds asymmetric noise, which moves labels
to a similar class, and 1{,}000 classes. Appendix~\ref{app:repro-main} describes every dataset.

\paragraph{Models.} We fine-tune ImageNet-pretrained ViT-S/16 \citep{dosovitskiy2021vit,steiner2022augreg} and
ConvNeXt-T \citep{liu2022convnext} from \texttt{timm} \citep{wightman2019timm} with plain cross-entropy for 40 epochs
(AdamW, batch size 128, cosine decay, resolution 224), at the learning rate selected by a search on a noisy 10\% split
of the training set (Appendix~\ref{app:repro}). Training runs well past the point where test accuracy peaks, so the
wrong labels are fitted. We keep the model after 2, 5, 10, 20 and 40 epochs and audit each one. Every setting is run
five times, with five consecutive training seeds.

\paragraph{Audit and compared scores.} Unless stated otherwise the audit covers 5{,}000 training examples drawn at
random, and the share of wrong labels among them is the truth, used for scoring only. We compare the audit with scores
computed from the same inputs: the loss, the feature margin, the label vote of the ten nearest neighbours in the
model's feature space, the gradient norm and the self-influence. TracIn, which needs the checkpoints of the training
run, and confident learning, which trains four further models, are reported with these extra inputs stated. Detectors
that do not read the trained model, such as SimiFeat \citep{zhu2022simifeat}, address a different setting and are not
compared. \emph{Identification}, which labels are wrong, is measured by the AUC of a score and by the precision and
recall of a flag. \emph{Estimation}, how many labels are wrong, is measured by the difference between the reading and
the true share.

\subsection{Find: identifying wrong labels in a fitted model}
\label{sec:survive}

\begin{figure*}[t]
\centering
\begin{tikzpicture}
\begin{groupplot}[group style={group size=4 by 1, horizontal sep=1.0cm}, width=0.275\textwidth, height=4.6cm,
  xlabel={wrong labels fitted (\%)}, ymin=0.70, ymax=1.0, xmin=0, xmax=102, grid=major, grid style={gray!25},
  legend style={font=\scriptsize, draw=none, fill=none}, tick label style={font=\scriptsize}, label style={font=\scriptsize},
  title style={font=\scriptsize}]
\nextgroupplot[title={ViT-S, C10N worst}, ylabel={AUC, wrong vs.\ right labels}, legend pos=south west]
\addplot[blue!75!black, thick, mark=*, mark size=1.3pt] coordinates {(7.8,0.983) (11.2,0.982) (33.4,0.972) (88.6,0.963) (98.8,0.963)};
\addplot[orange!85!black, thick, mark=square*, mark size=1.3pt] coordinates {(7.8,0.990) (11.2,0.987) (33.4,0.970) (88.6,0.895) (98.8,0.844)};
\addplot[gray!70!black, thick, mark=triangle*, mark size=1.5pt] coordinates {(7.8,0.981) (11.2,0.983) (33.4,0.925) (88.6,0.770) (98.8,0.759)};
\legend{allocation, feature, loss}
\nextgroupplot[title={ConvNeXt-T, C10N worst}]
\addplot[blue!75!black, thick, mark=*, mark size=1.3pt] coordinates {(8.5,0.976) (16.4,0.979) (73.9,0.946) (99.1,0.942) (100.0,0.940)};
\addplot[orange!85!black, thick, mark=square*, mark size=1.3pt] coordinates {(8.5,0.989) (16.4,0.983) (73.9,0.908) (99.1,0.807) (100.0,0.790)};
\addplot[gray!70!black, thick, mark=triangle*, mark size=1.5pt] coordinates {(8.5,0.982) (16.4,0.976) (73.9,0.785) (99.1,0.726) (100.0,0.737)};
\nextgroupplot[title={ViT-S, C100N}]
\addplot[blue!75!black, thick, mark=*, mark size=1.3pt] coordinates {(15.1,0.919) (26.3,0.930) (69.4,0.921) (98.8,0.915) (99.9,0.912)};
\addplot[orange!85!black, thick, mark=square*, mark size=1.3pt] coordinates {(15.1,0.953) (26.3,0.953) (69.4,0.920) (98.8,0.884) (99.9,0.877)};
\addplot[gray!70!black, thick, mark=triangle*, mark size=1.5pt] coordinates {(15.1,0.935) (26.3,0.935) (69.4,0.866) (98.8,0.796) (99.9,0.800)};
\nextgroupplot[title={ConvNeXt-T, C100N}]
\addplot[blue!75!black, thick, mark=*, mark size=1.3pt] coordinates {(15.6,0.806) (23.8,0.888) (62.7,0.881) (99.1,0.872) (100.0,0.863)};
\addplot[orange!85!black, thick, mark=square*, mark size=1.3pt] coordinates {(15.6,0.938) (23.8,0.953) (62.7,0.914) (99.1,0.869) (100.0,0.856)};
\addplot[gray!70!black, thick, mark=triangle*, mark size=1.5pt] coordinates {(15.6,0.920) (23.8,0.939) (62.7,0.876) (99.1,0.777) (100.0,0.776)};
\end{groupplot}
\end{tikzpicture}
\caption{Separation of wrong from right labels as the wrong labels are fitted. Each point is the model kept after 2, 5,
10, 20 or 40 epochs of cross-entropy training, audited on the whole training set, mean of five runs (standard deviation
at most 0.02). The horizontal axis is the share of wrong labels that the model predicts.}
\label{fig:auc}
\end{figure*}
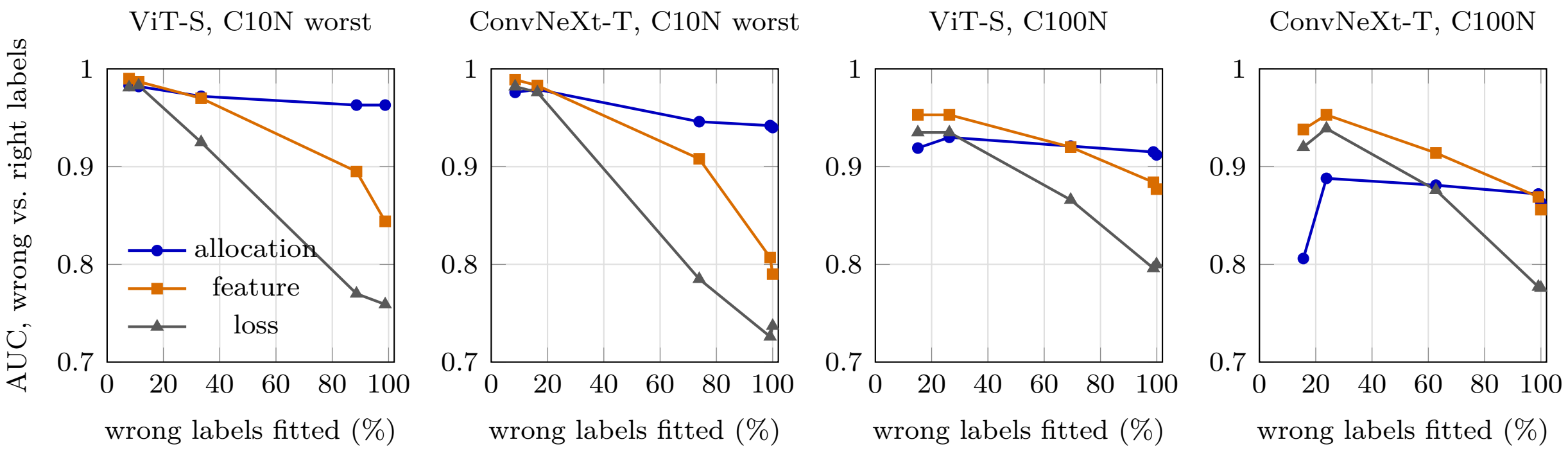

Figure~\ref{fig:auc} follows the area under the ROC curve (AUC) of three scores for separating wrong from right labels as
the wrong labels are fitted: the allocation margin of Eq.~\eqref{eq:margin}, the same margin computed on the classifier's
input features, and the loss. While the model has fitted few wrong labels, the three scores are close, and the feature
margin is the best of them. As the wrong labels are fitted, the loss and the feature margin lose most of their
separation. At the last kept model 99--100\% of the wrong labels are predicted, and the allocation margin keeps an AUC of
0.94--0.96 on CIFAR-10N and 0.86--0.91 on CIFAR-100N. Appendix~\ref{app:figures} shows the distributions of the scores.

Over the 22 noisy settings the label-swap score is the highest of the scores that read only the final model in every
setting, with a mean AUC of 0.949 against 0.73 to 0.89 (Table~\ref{tab:matrix} in Appendix~\ref{app:matrix}). A paired
comparison over the settings places it above each of them (Wilcoxon signed-rank test, $p \le 4.8 \times 10^{-7}$).
TracIn, which sums the self-influence over the checkpoints kept during training, reaches a mean of 0.935 and does not
differ from the audit ($p = 0.33$). Given the same history, the checkpoint after two epochs is easier still to audit,
because the model has not yet fitted the wrong labels (Appendix~\ref{app:flags}). The allocation is therefore the score
to use when only the final model is available, which is the case this paper addresses.

\paragraph{Flags and proposed labels.} At their natural cut, a negative margin and disagreement between prediction and
label, the feature margin and the prediction recover at most 7\% of the wrong labels of a fitted model, because the
model now treats them as correct. The label-swap rule recovers 31--72\% at a precision of 0.94--0.98
(Appendix~\ref{app:flags}). Its proposed label $a_i$ is the true class for 39 to 93\% of the wrong labels (median 73\%
over the 19 settings whose true classes are known), and the nearest other class in feature space for 0 to 8\%: fitting
moves the features of a wrong label away from its true class, while the allocation keeps pointing to it
(Figures~\ref{fig:nearest} and~\ref{fig:examples}).

\subsection{Count: a lower bound on the noise rate}
\label{sec:reading}

The reading is the share of flagged examples, so it is tied to the quality of the flags by an identity. With $\rho$ the
true share of wrong labels and $P$ and $R$ the precision and recall of the flags,
\begin{equation}
\text{reading} = \rho \, \frac{R}{P}.
\label{eq:identity}
\end{equation}
A good detector is therefore not automatically a good estimator. The reading is accurate when the flags miss about as
many wrong labels as they add right ones ($R \approx P$), and it is a lower bound when the flags are precise but
incomplete ($R < P$).

Before the wrong labels are fitted the reading follows the truth. After two epochs it lies within 1.5 percentage points
of the true share for ViT-S and within 1.7 for ConvNeXt-T at every noise level on CIFAR-10, from 0\% to 58\%. Once the
model predicts almost all of its training labels, the reading falls below the truth and becomes a lower bound. For ViT-S
on CIFAR-10 it reads 54--96\% of the true share, and on CIFAR-10, CIFAR-100 and Food-101 the flags keep a mean precision of
0.93 or higher,
so the examples counted are almost all wrong. On clean training sets the audit flags at most 1.0\% of the examples.
Appendix~\ref{app:reading} gives every setting.

The wrong labels that the audit misses are mostly those that are easy to confuse. On CIFAR-100N, 35\% of the wrong human
labels name a class of the same superclass as the true class, and the audit recovers 28\% of them against 54\% of the
errors across superclasses. On CIFAR-10H it recovers 11--18\% of the wrong labels on images where fewer than 90\% of the
annotators chose the true class, and 39\% on the others (Appendix~\ref{app:further}). With 1{,}000 classes it recovers
97\% of the labels moved to a random species and 49\% of those moved to a related one (Section~\ref{sec:cost}). The
reading therefore counts the clear-cut errors. A batch of labels whose reading exceeds an acceptance threshold carries
at least about that share of wrong labels, since nearly all of the flagged examples are wrong.

\paragraph{Readings from predictions and features.} Table~\ref{tab:readcmp} lists other threshold-free readings of the
same models: the share of training examples whose prediction disagrees with the label, whose feature margin is
negative, or whose ten nearest neighbours in feature space mostly carry another label. After two epochs the
disagreement and the feature margin read the noise rate as well as the audit, because the model has not yet fitted its
wrong labels. After 40 epochs the disagreement reads 0.0--1.9\% and the feature margin 0.4--5.9\% at true rates of
8--58\%, and the neighbour vote keeps at most 25\%, while the audit still reads 54--96\% of the true rate.

\begin{table*}[t]
\centering
\caption{Threshold-free readings of the noise rate (\%) from the same ViT-S models on the same 5{,}000 training
examples, mean of five runs. Disagreement: prediction $\neq$ label. Feature margin: Eq.~\eqref{eq:margin} on the
classifier's input features, below zero. Neighbour vote: most of the ten nearest training examples in feature space
carry another label.}
\label{tab:readcmp}
\small\setlength{\tabcolsep}{4pt}
\begin{tabular}{lcccccccc}
\toprule
 & \multicolumn{6}{c}{CIFAR-10} & \multicolumn{2}{c}{CIFAR-100} \\
\cmidrule(lr){2-7}\cmidrule(lr){8-9}
 & clean & aggr. & rand.\,1 & sym.\,20 & worst & sym.\,60 & clean & C100N \\
\midrule
Truth & 0.0 & 8.3 & 16.5 & 20.3 & 39.4 & 58.4 & 0.0 & 41.3 \\
\midrule
\multicolumn{9}{l}{\emph{After epoch 2}} \\
Disagreement & 1.1 & 9.1 & 17.7 & 22.0 & 39.5 & 61.0 & 8.8 & 42.0 \\
Feature margin & 1.1 & 9.1 & 17.3 & 21.9 & 39.2 & 59.7 & 9.0 & 42.7 \\
Neighbour vote & 0.9 & 9.0 & 17.6 & 22.3 & 45.0 & 84.9 & 10.0 & 50.8 \\
Label-swap (ours) & 1.3 & 9.2 & 17.4 & 21.8 & 38.8 & 59.2 & 9.0 & 40.3 \\
\midrule
\multicolumn{9}{l}{\emph{After epoch 40}} \\
Disagreement & 0.0 & 0.0 & 0.1 & 0.2 & 0.6 & 1.9 & 0.0 & 0.0 \\
Feature margin & 0.0 & 0.4 & 0.9 & 2.0 & 2.9 & 5.9 & 0.0 & 1.1 \\
Neighbour vote & 0.0 & 0.9 & 3.1 & 8.8 & 9.0 & 25.0 & 0.0 & 4.0 \\
Label-swap (ours) & 0.2 & 4.5 & 11.8 & 19.4 & 28.9 & 54.8 & 1.0 & 19.2 \\
\bottomrule
\end{tabular}
\end{table*}

\subsection{Comparison with confident learning}
\label{sec:cl}

Confident learning \citep{northcutt2021confident} answers both questions from out-of-sample predicted probabilities.
We run it with the reference implementation (\texttt{cleanlab}) on four-fold cross-fitting of the training set: each
fold is predicted by a ViT-S trained on the other three folds for 10 epochs, which takes four further trainings of
about six minutes each on CIFAR. Table~\ref{tab:cl-main} summarises the comparison with the label-swap audit of the
final model on the same 5{,}000 examples, and Appendix~\ref{app:cl} gives every setting.

\begin{table}[t]
\centering
\caption{Confident learning (CL) against the label-swap audit (LS) of the final model, ViT-S, mean over the 11 noisy
settings (five runs each, the same 5{,}000 training examples). Reading: share of examples flagged, relative to the true
share. Clean: share of examples flagged on clean labels. Every setting is in Table~\ref{tab:cl}.}
\label{tab:cl-main}
\small\setlength{\tabcolsep}{4pt}
\begin{tabular}{lcc}
\toprule
 & CL & LS \\
\midrule
Needs & 4 further trainings & final model \\
Time on CIFAR & about 23 min & 14 s \\
AUC & 0.969 & 0.958 \\
Precision of the flags & 0.85 & 0.94 \\
Reading / truth & 93\% & 74\% \\
Clean C10 / C100 / C10H (\%) & 0.3 / 4.6 / 0.5 & 0.2 / 1.0 / 0.2 \\
\bottomrule
\end{tabular}
\end{table}

Confident learning is given more than the audit: the training recipe, four further trainings and the predicted
probabilities of every training example. With these, it is not significantly better at separating wrong from right
labels (mean AUC 0.969 against 0.958, Wilcoxon signed-rank test over the 11 noisy settings, $p = 0.21$). It reads closer
to the true rate, 93\% of it on average against 74\% for the audit, but its flags are less precise (0.85 against 0.94,
and the audit is more precise in all 11 settings, $p = 9.8 \times 10^{-4}$), and on clean CIFAR-100 it flags 4.6\% of the
examples against 1.0\%. As a lower bound for accepting labels, the reading of the audit is therefore the safer one. Under
asymmetric noise the audit separates wrong from right labels better than confident learning in every run
(Section~\ref{sec:cost}).

\paragraph{Cost and accuracy.} The scores that come close to the audit or exceed it in some settings use more:
confident learning trains four further models, about 23 GPU-minutes for ViT-S against 14 seconds for the audit of
5{,}000 examples, and TracIn and the early-epoch scores need checkpoints from the training run. The scores that need
less, a forward pass for the loss and the feature-space scores, are less accurate after fitting
(Section~\ref{sec:survive}). In cost against accuracy the audit is therefore a point on the Pareto front: no score we
compare is both cheaper in what it needs and more accurate.

\subsection{Fix: retraining without the flagged examples}
\label{sec:retrain}

\begin{figure*}[t]
\centering
\includegraphics[width=\textwidth]{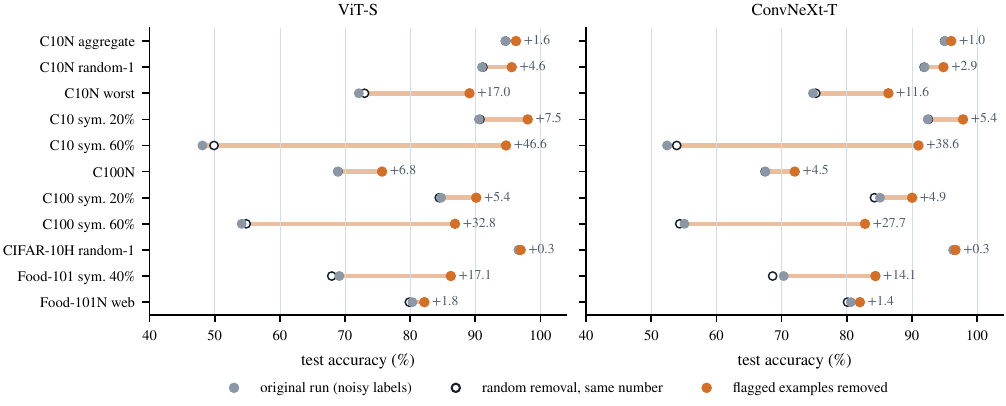}
\caption{Retraining without the flagged examples, mean of five runs: test accuracy of the original run on noisy
labels (grey), after removing the examples the audit flags (orange; difference to the original run on the right)
and after removing as many random examples (open). Same recipe and seed as the original run.}
\label{fig:retrain}
\end{figure*}

The flags can be acted on directly. For every noisy setting and run we audit the final model on its whole training
set, remove the flagged examples and train once more with the recipe and seed of the original run. As controls we
remove as many examples drawn at random, and we do the same on clean labels. Removing the flagged examples raises the
test accuracy in all 22 noisy settings, by 11.5 points on average and by 28 to 47 points with 60\% symmetric noise
(Wilcoxon signed-rank test over the settings, $p = 4.8 \times 10^{-7}$, Figure~\ref{fig:retrain}). Removing as many random
examples changes the accuracy by $-1.7$ to $+1.8$ points, so the gain comes from which examples are removed and not
from training on fewer of them. On clean labels the audit removes at most 1.2\% of the training set, and retraining
changes the accuracy by at most 0.13 points (Appendix~\ref{app:retrain}).

On the 11 noisy ViT-S settings we also remove, at the same budget, the examples with the highest loss under the final
model, and the label issues that confident learning finds, as many as it finds. Removing the audit's flags gives a test
accuracy of 90.2\% on average, against 85.8\% after removal by loss and 90.3\% after removal by confident learning,
which is level with it ($p = 0.70$) but removes 1.0 to 2.8 times as many examples and needs the four further
trainings. Of the examples the audit removes, 93.7\% carry wrong labels on average, against 75.3\% for removal by loss
and 84.6\% for confident learning (Appendix~\ref{app:cleaning}).

\subsection{Without shared data: auditing at the owner's site}
\label{sec:private}

\paragraph{What the owner of the data computes.} The split audit of Section~\ref{sec:method} asks the owner for
statistics computed from the monitored blocks, and cheaper inputs do not carry the signal once the labels are fitted.
Over the 11 noisy ViT-S settings, the same audit computed from the classifier head alone, that is from the final
features and the output probabilities, reaches an AUC of 0.957 after two epochs but 0.654 at the last kept model,
against 0.958 for the four monitored blocks, which are higher in all 11 settings ($p = 9.8 \times 10^{-4}$). Fitting
rewrites the final features more than the activations of the earlier blocks (Figure~\ref{fig:mechanism}).

\begin{figure}[t]
\centering
\includegraphics[width=3.25in]{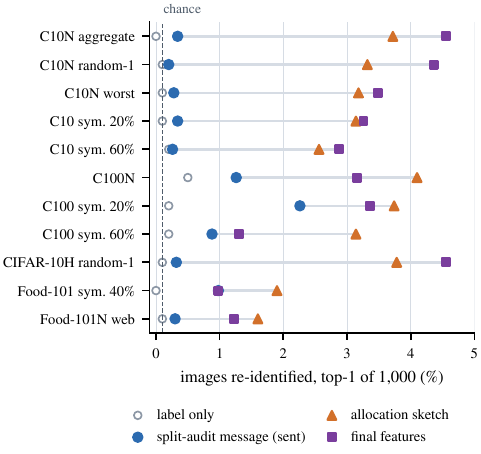}
\caption{Share of the owner's images that a decoder identifies among 1{,}000 from what it is given (ViT-S, final
model, mean of five runs; dashed line: chance). The split audit sends the message (blue). Reconstructions are shown in
Figure~\ref{fig:reconstruct}.}
\label{fig:reid}
\end{figure}

\paragraph{What the auditor can reconstruct.} The message protects the images well, even against an attacker stronger
than the auditor. Besides the trained model, the attacker holds the owner's class sums, which the split audit never
sends, and labelled images from the same distribution. It trains a decoder to map the message of an image back to the
image and applies it to the messages of the owner's 5{,}000 audited examples (Appendix~\ref{app:privacy}). From the
$2K + 4$ numbers the reconstruction is no closer to the image than the mean image of its class, with a PSNR of 12.4
against 12.6\,dB over the 11 noisy ViT-S settings, and the true image is the nearest of 1{,}000 candidates for 0.67\%
of the examples (Figure~\ref{fig:reid}). The same decoder re-identifies more than four times as many examples from the
final features (3.0\%) and from a 4{,}096-number sketch of the allocation itself (3.1\%), in every setting where the two
differ ($p \le 2.0 \times 10^{-3}$).

\subsection{Text classification}
\label{sec:text}

We fine-tune RoBERTa-base \citep{liu2019roberta} and DeBERTa-v3-base \citep{he2023debertav3} on the worst human labels
of NoisyAG-News \citep{noisyagnews} (37.8\% wrong) and on SST-2, QQP and MNLI of GLUE \citep{wang2019glue} with 40\%
symmetric noise, and Qwen3-1.7B-Base \citep{qwen3} on NoisyAG-News, five runs each, and audit the last four transformer
blocks of the final model (Appendix~\ref{app:repro-ext}). The label-swap score is the highest final-model score in all
nine settings, with a mean AUC of 0.723 against 0.53 to 0.58 for the others (Wilcoxon signed-rank test over the
settings, $p = 0.0039$ against each, the smallest value nine settings allow). TracIn over the kept checkpoints reaches
0.760 on average and does not differ significantly ($p = 0.12$). Once the wrong labels are fitted the reading is a
lower bound on text as well (Appendix~\ref{app:text}).

\subsection{Noise type, number of classes, ablations and cost}
\label{sec:cost}

\paragraph{Asymmetric noise.} Annotators confuse similar classes, and a label moved to a similar class is fitted with
confidence. We move 40\% of the labels of each affected class to a fixed partner class, along the standard pairs on
CIFAR-10 (20.1\% of the audited labels wrong) and to the next class of the same superclass on CIFAR-100 (39.9\%), with
ViT-S and five runs each. The label-swap score separates wrong from right labels with an AUC of 0.992 and 0.935,
against 0.956 and 0.796 for the best other score of the final model, the feature margin, 0.909 and 0.709 for TracIn,
and 0.919 and 0.691 for confident learning with its four further trainings. It is the highest of all of them in each
of the ten runs, and its flags are more precise than those of confident learning in each run (0.96 and 0.95 against
0.77 and 0.56, Appendix~\ref{app:asym}).

\paragraph{A backbone with batch normalisation.} With an ImageNet-pretrained ResNet-50
\citep{he2016resnet,wightman2021resnet}, trained on CIFAR-10N worst and CIFAR-100N with a learning-rate search of its own
and five runs, the label-swap score reaches 0.958 and 0.905, between ViT-S (0.965 and 0.912) and ConvNeXt-T (0.954 and
0.874), and it is the highest of the seven scores in each of the ten runs (Appendix~\ref{app:resnet}).

\paragraph{One thousand classes.} With many classes the class sums take more memory, and a wrong label has more classes
to point to. We draw 1{,}000 of the 10{,}000 species of iNaturalist 2021 \citep{vanhorn2021inat} and train ViT-S on their
50{,}000 training images with 40\% symmetric noise or with 40\% asymmetric noise to the next drawn species of the same
supercategory, five runs each, and audit the whole training set. The final models have fitted all wrong labels. The
label-swap score reaches an AUC of 0.991 and 0.852, against 0.971 and 0.778 for the best other score, and it is the
highest of the seven scores in each of the ten runs. Its flags have a precision of 0.916 and 0.837 and recover 97\% and
49\% of the wrong labels (Appendix~\ref{app:inat}).

\paragraph{Ablations.} Table~\ref{tab:ablation} in Appendix~\ref{app:ablation} answers one question per row on
CIFAR-10N worst, where the audit reaches an AUC of 0.965 at the final model. Both steps of the allocation are needed:
the raw magnitudes $|g_i|$ give 0.784 and the signed gradient normalised in the same way 0.765. The last block alone
gives 0.919, and four monitored blocks recover the signal that fitting removes from it. Full FP32 precision and batch
sizes of 64 or 256 leave the AUC within 0.001. About 50 audited examples per class suffice. A random 10\% of the
monitored coordinates changes the AUC by at most 0.002 for ViT-S, ResNet-50 and 1{,}000 classes, and it cuts the memory
of the class means by a factor of ten.

\paragraph{Cost.} The audit grows linearly with the number of examples. For 5{,}000 examples the label-swap audit takes
14 to 27 seconds on one RTX 4090, about one hundredth of the time it took to train the model, and with 1{,}000 classes
the audit of 50{,}000 examples takes 5 minutes (Appendix~\ref{app:cost}).

\section{Conclusion}

A model that has fitted its wrong labels still records them in where its per-example gradients point. Normalised into
an allocation over parameters, the gradient of a wrongly labelled example resembles its true class, and the label-swap
audit turns this into a threshold-free decision and a reading that is a lower bound on the noise rate. Across human,
web and synthetic noise, three pretrained image backbones, 1{,}000 classes and three language models, it is the most
accurate of the scores that read only the final model, and it is level with TracIn, which needs the checkpoints of
training. Removing its flags and training again raises the test accuracy by up to 47 points, level with the cleaning of
confident learning without its further trainings. Because the decision needs only a few numbers per example, which the
owner of the data computes and from which a decoder recovers no more than the mean image of the class, labels can be
audited where the data cannot be shared. A model can thus be audited in seconds as it is handed over, when labels are
accepted from annotators, when a trained model is taken over, or when examples are chosen for re-annotation.

\section{Scope}

The experiments cover image classification with two pretrained backbones on human, web and symmetric noise, with one
of them also on asymmetric noise and on 1{,}000 classes, a third backbone with batch normalisation on two sets of human
labels, and text classification with two encoders and a decoder language model. They compare the audit with
final-model scores, TracIn and confident learning on the same examples, retrain after removal, attack the message of
the split audit with a trained decoder, and vary the compared quantity, the monitored layers and coordinates, the
training precision, the batch size and the number of audited examples.

\appendix
\FloatBarrier
\section{Algorithm, derivations and additional figures}
\label{app:method}

\subsection{The split audit: sufficient statistics and algorithm}
\label{app:algorithm}

\paragraph{The statistics of the split audit suffice.} Write $d_{ic}(y) = \langle q_i(y), S_c \rangle$ and $n_c$ for
the number of audited examples with observed label $c$. A class mean is its class sum divided by $n_c$, so the cosine
with the mean equals the cosine with the sum. For a class sum that does not contain the row,
\begin{equation}
\cos\bigl(q_i(y), \mu_c\bigr) = \frac{d_{ic}(y)}{\lVert q_i(y) \rVert \, \lVert S_c \rVert},
\end{equation}
and for the observed label and its own class, with the example left out,
\begin{equation}
\cos\bigl(q_i(y_i), \mu_{y_i}\bigr) = \frac{d_{i y_i}(y_i) - \lVert q_i(y_i) \rVert^2}
{\lVert q_i(y_i) \rVert \sqrt{\lVert S_{y_i} \rVert^2 - 2 d_{i y_i}(y_i) + \lVert q_i(y_i) \rVert^2}}.
\end{equation}
The margins of Eq.~\eqref{eq:margin} under $y_i$ and under $a_i$ and the flag of Eq.~\eqref{eq:swap} are therefore
functions of the $2K$ inner products $d_{ic}(y_i)$ and $d_{ic}(a_i)$, the two squared norms and the two labels of the
example, together with $\lVert S_c \rVert$ and $n_c$ of every class. Algorithm~\ref{alg:split} lists the steps.
The owner makes one pass over the audited examples to form the class sums and one further pass per label considered,
each a forward and a backward computation of the frozen model. The class sums occupy one vector per class of the size
of the monitored parameters, and a random tenth of the coordinates is enough (Appendix~\ref{app:ablation}). Instead
of these statistics the owner could send the inputs and the returned errors of the monitored projections, from which
the auditor forms the allocations itself. The statistics above disclose less. For CIFAR-10 the message is 24 numbers per
example, against the 3{,}072 values of a $32 \times 32$ colour image.

\begin{algorithm}[t]
\caption{The split label-swap audit.}
\label{alg:split}
\begin{algorithmic}[1]
\STATE \textbf{Owner} (holds the trained model, the examples $x_i$ and their observed labels $y_i$):
\STATE first pass: compute $q_i(y_i)$ for every example and accumulate $S_c = \sum_{i: y_i = c} q_i(y_i)$ and $n_c$
\FOR{every example $i$ (second pass)}
\STATE compute $q_i(y_i)$, the inner products $d_{ic}(y_i)$ for all $c$ and $\lVert q_i(y_i) \rVert^2$
\STATE $a_i \leftarrow \arg\max_{c \neq y_i} \cos(q_i(y_i), \mu_c)$
\STATE compute $q_i(a_i)$ with $a_i$ as the label, the inner products $d_{ic}(a_i)$ for all $c$ and $\lVert q_i(a_i) \rVert^2$
\STATE send $d_{i\cdot}(y_i)$, $d_{i\cdot}(a_i)$, $\lVert q_i(y_i) \rVert^2$, $\lVert q_i(a_i) \rVert^2$, $y_i$, $a_i$ \quad ($2K + 4$ numbers)
\ENDFOR
\STATE send $\lVert S_c \rVert$ and $n_c$ for every class
\STATE \textbf{Auditor} (holds only what was sent):
\STATE compute $m_i(y_i)$ and $m_i(a_i)$ from the cosines above
\STATE flag example $i$ when $m_i(a_i) > m_i(y_i)$ and propose $a_i$ as its label
\STATE reading $\leftarrow$ share of flagged examples
\end{algorithmic}
\end{algorithm}

\subsection{Per-example gradients and class means}
\label{app:impl}

\paragraph{Per-example gradients.} For a linear layer $y = xW^\top + b$ applied at every token or spatial position $t$
of example $i$, the per-example gradient is $\sum_t \delta_{it}^\top x_{it}$ for the weights and $\sum_t \delta_{it}$
for the bias, where $\delta_{it}$ is the gradient of the example's loss with respect to $y_{it}$. A $1\times1$
convolution is such a layer. The model is in evaluation mode, in which the examples of a batch do not interact: a
batch-normalisation layer then applies its stored statistics to every example. One backward pass of the summed loss
therefore yields every $\delta_{it}$, and forward and backward hooks collect the inputs and output gradients. The forward and backward
computations run under bfloat16 autocast with float32 weights. The per-example rows are formed in float32 and all inner
products are accumulated in float64.

\paragraph{Class means and leave-one-out.} The audit keeps the class sums $S_c = \sum_{j: y_j = c} q_j(y_j)$, their
squared norms and the counts. For example $i$ the cosine with its own class uses $S_{y_i} - q_i(y_i)$, so no example is
compared with a mean that contains itself. The alternative label's allocation $q_i(a_i)$ is not part of any class sum.

\paragraph{What the allocation records.} What the allocation records is set by two factors. Each monitored projection is a weight matrix $W$ applied at every
token or spatial position, so the gradient of an example is $\sum_t \delta_t a_t^\top$, with $a_t$ the input of the
projection at position $t$ and $\delta_t$ the error signal returned to its output. The activations $a_t$ carry the
content of the input, and the direction of the errors depends on the label. A cat labelled as a dog keeps the activations of a cat and receives the error pattern of the label
\emph{dog}. Its allocation therefore resembles neither the dogs, whose activations differ, nor the cats, whose errors
differ, and relabelling it as a cat restores both factors. The label-swap test measures exactly this change. The
activations of earlier monitored blocks are altered less by fitting than the final features, which agrees with the
ablation in Section~\ref{sec:cost}: monitoring only the last block loses part of the signal after fitting, and
monitoring four blocks recovers it. Figure~\ref{fig:mechanism} sketches the argument. The second pass of the label-swap test is unaffected by fitting as well. For
an example fitted to its observed label $y_i$, $p_{y_i} \to 1$ gives $r_{y_i} \to 1$ under an alternative label $a$, so
$q_i(a)$ tends to the allocation of $\nabla_\theta (f_a - f_{y_i})$, a gradient whose size does not vanish.

\subsection{Additional figures}
\label{app:figures}

\begin{figure*}[t]
\centering
\includegraphics[width=\textwidth]{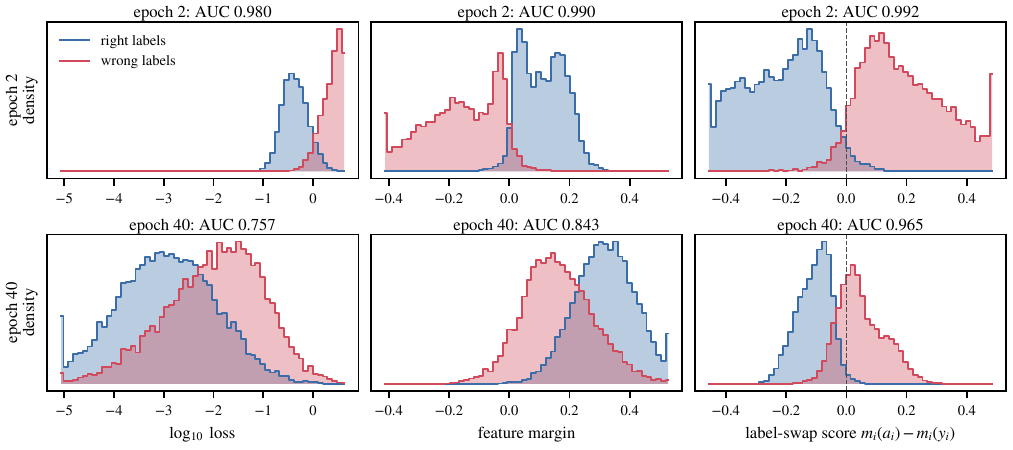}
\caption{Scores of right and wrong labels on CIFAR-10N worst (ViT-S, five runs, 5{,}000 training examples each), after
epoch 2 (top) and after epoch 40 (bottom), when 99\% of the wrong labels are fitted. The label-swap score is
$m_i(a_i) - m_i(y_i)$; an example is flagged when it is positive (dashed line). AUC: mean over the runs.}
\label{fig:dist}
\end{figure*}

\paragraph{The distributions behind the AUC.} Figure~\ref{fig:dist} shows the scores themselves on CIFAR-10N worst.
After two epochs the loss, the feature margin and the label-swap score of Eq.~\eqref{eq:swap} all place the wrong
labels apart from the right ones. After forty epochs the loss and the feature margin of the wrong labels have moved onto
those of the right labels, while the label-swap scores of the wrong labels stay above zero for most of them and those
of the right labels stay below it.

\begin{figure}[t]
\centering
\includegraphics[width=0.62\linewidth]{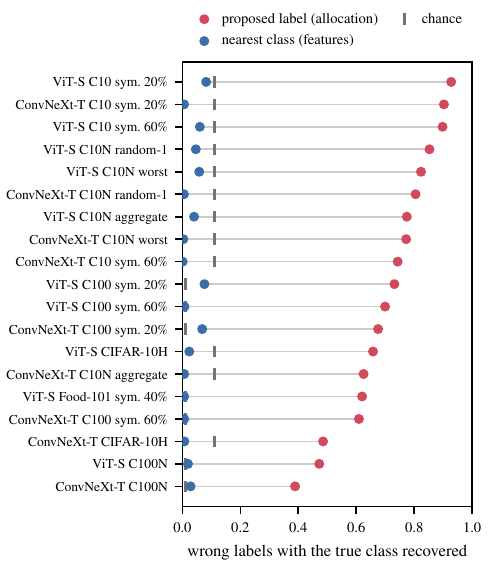}
\caption{Share of wrong labels whose true class is recovered at the last kept model: the audit's proposed label
(allocation) and the nearest other class in feature space; tick: chance, $1/(C-1)$. Mean of five runs; Food-101N is left
out because the true class of a wrong web label is unknown.}
\label{fig:nearest}
\end{figure}

\begin{figure*}[t]
\centering
\includegraphics[width=\textwidth]{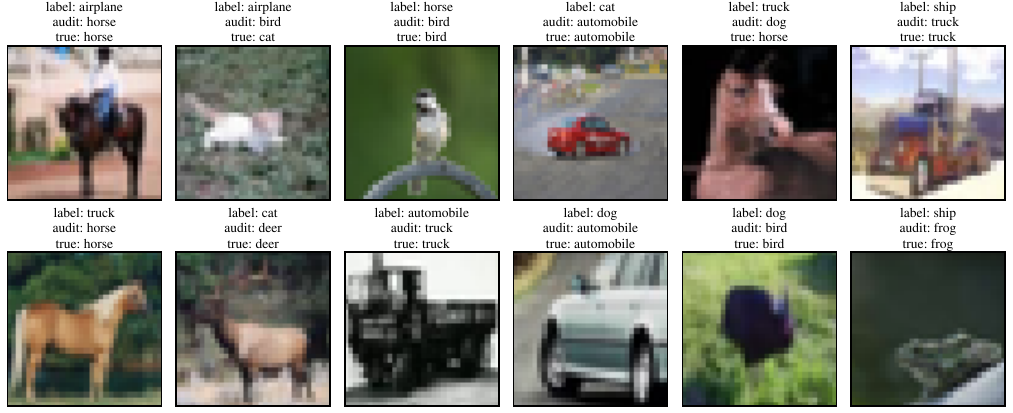}
\caption{The first twelve flagged training examples in index order (ViT-S, CIFAR-10N worst, one run, epoch 40), with
the observed label, the label proposed by the audit and the true label. No examples were selected.}
\label{fig:examples}
\end{figure*}

\paragraph{Where wrong labels point.} The proposed label $a_i$ of a wrong label is usually its true class
(Figure~\ref{fig:nearest}). Over the 19 noisy settings whose true classes are known, the proposed label of a wrong
label at the last kept model is the true class for 39 to 93\% of the wrong labels (median 73\%). The nearest other
class in feature space is the true class for 0 to 8\%, below chance in every CIFAR-10 setting: fitting
moves the features of a wrong label away from its true class, while the allocation keeps pointing to it.
Figure~\ref{fig:examples} shows the first twelve flagged examples of one run without selection: all twelve labels are
wrong, and for ten of them the proposed label is the true class.

\subsection{Reconstructions of the inversion attack}
\label{app:reconstruct}

\begin{figure*}[t]
\centering
\includegraphics[width=\textwidth]{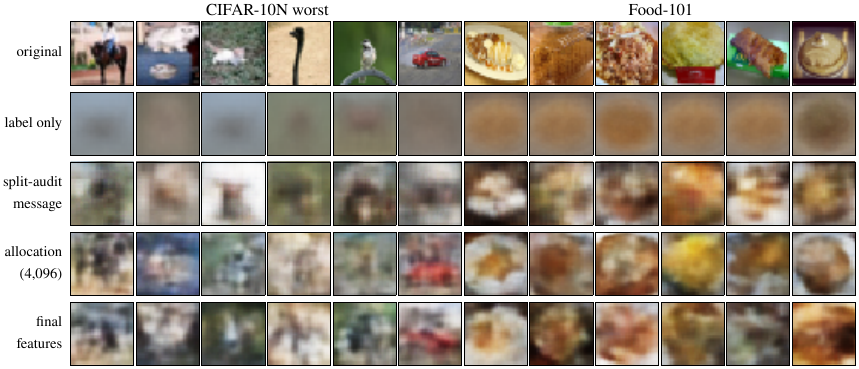}
\caption{Reconstructions of the inversion attack (ViT-S, final model of one run), for the first six audited examples
of CIFAR-10N worst and Food-101 in index order, with no selection. Rows: the original image, the label alone (mean
image of the class), and the decoder's output from the split-audit message, from a sketch of the allocation and from
the final features. The message yields images typical of the class rather than the original; over the 11 settings
their PSNR is 12.4\,dB against 12.6\,dB for the class-mean image (Table~\ref{tab:privacy}).}
\label{fig:reconstruct}
\end{figure*}

Figure~\ref{fig:reconstruct} shows what the inversion attack of Section~\ref{sec:private} reconstructs. Its results are
in Appendix~\ref{app:privacy} and its setup in Appendix~\ref{app:privacy-setup}.

\FloatBarrier
\section{Main experiments: every setting and every run}
\label{app:main-results}

\subsection{Final-model scores in every setting}
\label{app:matrix}

Table~\ref{tab:matrix} lists the AUC for wrong vs.\ right labels of every score at the last kept model, on the same
5{,}000 training examples of each run. TracIn sums the self-influence over all kept checkpoints and is the only column
that uses the training history. Over the 22 rows the label-swap score has the highest mean AUC of the final-model
scores and the highest AUC in all 22 rows; a Wilcoxon signed-rank test over the rows places it above each final-model score
($p \le 4.8 \times 10^{-7}$) and level with TracIn ($p = 0.33$).

\begin{table}[h]
\centering
\caption{AUC at the last kept model (epoch 40; CIFAR-10H epoch 80), mean of five runs. Bold: the highest
final-model score of the row. LS: label-swap (ours); FM:
feature margin; NV: neighbour vote (kNN-10); GN: gradient norm; SI: self-influence; TracIn: summed over the kept
checkpoints.}
\label{tab:matrix}
\small\setlength{\tabcolsep}{3.5pt}
\begin{tabular}{llccccccc}
\toprule
 & & \multicolumn{6}{c}{Final model only} & History \\
\cmidrule(lr){3-8}\cmidrule(lr){9-9}
Model & Labels & LS & Loss & FM & NV & GN & SI & TracIn \\
\midrule
ViT-S & C10N aggregate & \textbf{0.958} & 0.880 & 0.912 & 0.623 & 0.882 & 0.881 & 0.987 \\
 & C10N random-1 & \textbf{0.972} & 0.842 & 0.895 & 0.719 & 0.837 & 0.837 & 0.985 \\
 & C10N worst & \textbf{0.965} & 0.757 & 0.843 & 0.758 & 0.748 & 0.747 & 0.934 \\
 & C10 sym.\ 20\% & \textbf{0.995} & 0.929 & 0.966 & 0.898 & 0.913 & 0.912 & 0.985 \\
 & C10 sym.\ 60\% & \textbf{0.995} & 0.757 & 0.865 & 0.885 & 0.737 & 0.734 & 0.842 \\
 & C100N & \textbf{0.912} & 0.799 & 0.869 & 0.685 & 0.799 & 0.797 & 0.920 \\
 & C100 sym.\ 20\% & \textbf{0.993} & 0.960 & 0.982 & 0.897 & 0.949 & 0.947 & 0.962 \\
 & C100 sym.\ 60\% & \textbf{0.993} & 0.867 & 0.935 & 0.897 & 0.844 & 0.840 & 0.918 \\
 & CIFAR-10H random-1 & \textbf{0.900} & 0.830 & 0.863 & 0.562 & 0.836 & 0.836 & 0.962 \\
 & Food-101 sym.\ 40\% & \textbf{0.984} & 0.859 & 0.926 & 0.881 & 0.834 & 0.832 & 0.930 \\
 & Food-101N web & \textbf{0.876} & 0.816 & 0.844 & 0.631 & 0.810 & 0.807 & 0.854 \\
\midrule
ConvNeXt-T & C10N aggregate & \textbf{0.931} & 0.849 & 0.879 & 0.570 & 0.851 & 0.848 & 0.982 \\
 & C10N random-1 & \textbf{0.955} & 0.820 & 0.860 & 0.619 & 0.815 & 0.813 & 0.984 \\
 & C10N worst & \textbf{0.954} & 0.741 & 0.794 & 0.607 & 0.735 & 0.733 & 0.948 \\
 & C10 sym.\ 20\% & \textbf{0.993} & 0.925 & 0.956 & 0.773 & 0.913 & 0.912 & 0.986 \\
 & C10 sym.\ 60\% & \textbf{0.987} & 0.742 & 0.787 & 0.695 & 0.727 & 0.725 & 0.845 \\
 & C100N & \textbf{0.874} & 0.777 & 0.849 & 0.663 & 0.781 & 0.777 & 0.918 \\
 & C100 sym.\ 20\% & \textbf{0.991} & 0.963 & 0.984 & 0.891 & 0.954 & 0.952 & 0.961 \\
 & C100 sym.\ 60\% & \textbf{0.986} & 0.876 & 0.938 & 0.902 & 0.853 & 0.849 & 0.897 \\
 & CIFAR-10H random-1 & \textbf{0.839} & 0.799 & 0.832 & 0.540 & 0.799 & 0.794 & 0.968 \\
 & Food-101 sym.\ 40\% & \textbf{0.972} & 0.854 & 0.921 & 0.848 & 0.837 & 0.834 & 0.919 \\
 & Food-101N web & \textbf{0.860} & 0.801 & 0.829 & 0.588 & 0.804 & 0.798 & 0.887 \\
\midrule
\multicolumn{2}{l}{Mean of 22 settings} & \textbf{0.949} & 0.838 & 0.888 & 0.733 & 0.830 & 0.827 & 0.935 \\
\bottomrule
\end{tabular}
\end{table}

\subsection{Flags and gradient-based scores}
\label{app:flags}

The difference matters for decisions. Table~\ref{tab:flags} lists what each score flags at the last kept model when it
is given its natural cut: a negative margin, and disagreement between prediction and label. The feature margin and the
prediction recover at most 7\% of the wrong labels there, because the model now treats them as correct. The allocation
margin recovers 21--64\% with a precision of 0.95--0.99, and the label-swap rule raises this to 31--72\% with a
precision of 0.94--0.98.

\begin{table*}[t]
\centering
\caption{Flags at the last kept model (99--100\% of wrong labels fitted): share of the wrong labels that are flagged,
with the precision of the flags in parentheses for the two allocation rules, mean of five runs. Label-swap: Eq.~\eqref{eq:swap}, 5{,}000 examples. Other rows: whole training set.}
\label{tab:flags}
\small\setlength{\tabcolsep}{3.5pt}
\begin{tabular}{lcccc}
\toprule
 & \multicolumn{2}{c}{CIFAR-10N worst} & \multicolumn{2}{c}{CIFAR-100N} \\
\cmidrule(lr){2-3}\cmidrule(lr){4-5}
Flag & ViT-S & ConvNeXt-T & ViT-S & ConvNeXt-T \\
\midrule
Prediction $\neq$ label & 0.01 & 0.00 & 0.00 & 0.00 \\
Feature margin $<0$ & 0.07 & 0.00 & 0.03 & 0.04 \\
Allocation margin $<0$ & 0.64 (0.98) & 0.49 (0.99) & 0.40 (0.97) & 0.21 (0.95) \\
Label-swap & 0.72 (0.97) & 0.61 (0.98) & 0.45 (0.96) & 0.31 (0.94) \\
\bottomrule
\end{tabular}
\end{table*}

\paragraph{Gradient-based scores.} Table~\ref{tab:grad} compares the audit with scores that are also computed from
per-example gradients, on the same monitored layers and the same rows. With the final model alone, the gradient norm
(as in GraNd, \citealp{paul2021el2n}) and the self-influence (TracIn with one checkpoint, \citealp{pruthi2020tracin}) lose
their separation together with the loss. TracIn summed over the five kept checkpoints stays strong except at 60\%
symmetric noise (0.84), and it is the best score for ConvNeXt-T on CIFAR-100N, but it needs the training history. Given
the same history, the earliest kept checkpoint is easier to audit than the final model: its label-swap audit reaches a
similar or higher AUC than TracIn in every setting, and its loss comes close, because the model has not yet fitted the
wrong labels (Figure~\ref{fig:auc}). The allocation is therefore the score to use when only the final model is
available, which is the case this paper addresses.

\begin{table*}[t]
\centering
\caption{AUC for wrong vs.\ right labels of gradient-based scores on the same 5{,}000 training examples, mean of five
runs. Aggr., rand.\,1 and worst are the human labels of CIFAR-10N, sym.\,20 and sym.\,60 are CIFAR-10 with 20\% and
60\% symmetric noise. The upper block reads only the final model (epoch 40). The lower block also uses earlier
checkpoints of the same training run: TracIn sums the self-influence over the five kept checkpoints, and the last two
rows read the checkpoint after epoch 2.}
\label{tab:grad}
\small\setlength{\tabcolsep}{4pt}
\begin{tabular}{lcccccccc}
\toprule
 & \multicolumn{6}{c}{ViT-S/16} & \multicolumn{2}{c}{ConvNeXt-T} \\
\cmidrule(lr){2-7}\cmidrule(lr){8-9}
Score & aggr. & rand.\,1 & worst & sym.\,20 & sym.\,60 & C100N & worst & C100N \\
\midrule
\multicolumn{9}{l}{\emph{Final model only}} \\
Loss & 0.880 & 0.842 & 0.757 & 0.929 & 0.757 & 0.799 & 0.741 & 0.777 \\
Gradient norm & 0.882 & 0.837 & 0.748 & 0.913 & 0.737 & 0.799 & 0.735 & 0.781 \\
Self-influence & 0.881 & 0.837 & 0.747 & 0.912 & 0.734 & 0.797 & 0.733 & 0.777 \\
Label-swap (ours) & 0.958 & 0.972 & 0.965 & 0.995 & 0.995 & 0.912 & 0.954 & 0.874 \\
\midrule
\multicolumn{9}{l}{\emph{Training history}} \\
TracIn, 5 checkpoints & 0.987 & 0.985 & 0.934 & 0.985 & 0.842 & 0.920 & 0.948 & 0.918 \\
Loss, epoch 2 & 0.990 & 0.993 & 0.980 & 0.999 & 0.992 & 0.932 & 0.981 & 0.917 \\
Label-swap, epoch 2 & 0.991 & 0.996 & 0.992 & 0.999 & 0.998 & 0.944 & 0.989 & 0.910 \\
\bottomrule
\end{tabular}
\end{table*}

\subsection{Noise-rate reading in every setting}
\label{app:reading}

Table~\ref{tab:reading} compares the reading of the label-swap audit with the true share of wrong labels. For a model
trained for two epochs the reading follows the truth at every noise level, human or synthetic, from 0\% to 58\%, with
an error of at most 1.5 percentage points for ViT-S and 1.7 for ConvNeXt-T on CIFAR-10. Its standard deviation across
the five training runs is at most 0.7 points. On CIFAR-100 the early reading lies above the truth under symmetric noise
(27.6\% and 29.6\% for 19.7\%, 66.6\% and 67.0\% for 60.4\%, ViT-S and ConvNeXt-T), for the same reason as on clean
CIFAR-100 (Appendix~\ref{app:further}). On Food-101 and Food-101N the early reading lies above the truth as well (52.5--53.5\% for 39.8\%
and 33.3--36.1\% for 19.2\%), at precisions of 0.74 and 0.44--0.45: after two epochs the model predicts only 47--48\% and
64--65\% of its training labels, and the flags include right labels it has not yet learned.

After 40 epochs, when the model predicts almost all of its training labels, the reading falls below the truth. For
symmetric noise it stays close, at 78--98\% of the true share for both backbones on CIFAR-10 and CIFAR-100 and at 80--96\%
on Food-101. For human noise it reads 43--73\% of the true share on CIFAR-10 (54--73\% for ViT-S) and 33--46\% on
CIFAR-100N, so the reading is a lower bound. On CIFAR-10, CIFAR-100 and Food-101 the flags keep a mean precision of 0.93 or higher, so the
examples counted are almost all wrong. For the web labels of Food-101N the audit reads 13.1\% (ViT-S) and 9.3\% (ConvNeXt-T) at
a true 19.2\%, with precisions of 0.74 and 0.79, so the reading is a lower bound there as well. Which of the two regimes a model is in shows in its training accuracy on the observed labels, which
the audit reports alongside the reading.

Among the other threshold-free readings of Table~\ref{tab:readcmp}, the disagreement and the feature margin read the
noise rate after two epochs within 2.6 points on CIFAR-10 and 1.4 points on CIFAR-100N, and the neighbour vote
over-reads at high noise. The early reading is therefore a check that the audit is calibrated, while the reading of a
fitted model, where the readings from predictions and features vanish, is the case the audit is built for.

\begin{table}[t]
\centering
\caption{Noise-rate reading (\%) of the label-swap audit on 5{,}000 training examples against the true share of wrong
labels among them. Mean of five runs; standard deviation across runs at most 0.8 points.}
\label{tab:reading}
\small\setlength{\tabcolsep}{2.5pt}
\begin{tabular}{llcccc}
\toprule
 & & & \multicolumn{3}{c}{Reading after epoch} \\
\cmidrule(lr){4-6}
Model & Labels & Truth & 2 & 10 & 40 \\
\midrule
ViT-S & C10 clean & 0.0 & 1.3 & 0.6 & 0.2 \\
 & C10N aggregate & 8.3 & 9.2 & 5.7 & 4.5 \\
 & C10N random-1 & 16.5 & 17.4 & 13.7 & 11.8 \\
 & C10 sym.\ 20\% & 20.3 & 21.8 & 20.0 & 19.4 \\
 & C10N worst & 39.4 & 38.8 & 32.8 & 28.9 \\
 & C10 sym.\ 60\% & 58.4 & 59.2 & 56.8 & 54.8 \\
\midrule
ConvNeXt-T & C10 clean & 0.0 & 1.7 & 0.4 & 0.0 \\
 & C10N aggregate & 8.3 & 9.2 & 4.3 & 3.6 \\
 & C10N random-1 & 16.5 & 17.1 & 10.8 & 9.5 \\
 & C10 sym.\ 20\% & 20.3 & 22.0 & 18.9 & 18.7 \\
 & C10N worst & 39.4 & 38.0 & 25.4 & 24.5 \\
 & C10 sym.\ 60\% & 58.4 & 59.3 & 49.3 & 50.0 \\
\midrule
ViT-S & C100 clean & 0.0 & 9.0 & 3.1 & 1.0 \\
 & C100 sym.\ 20\% & 19.7 & 27.6 & 20.2 & 19.2 \\
 & C100N & 41.3 & 40.3 & 25.7 & 19.2 \\
 & C100 sym.\ 60\% & 60.4 & 66.6 & 58.2 & 56.2 \\
ConvNeXt-T & C100 clean & 0.0 & 10.0 & 2.9 & 1.0 \\
 & C100 sym.\ 20\% & 19.7 & 29.6 & 18.9 & 18.2 \\
 & C100N & 41.3 & 39.5 & 23.4 & 13.6 \\
 & C100 sym.\ 60\% & 60.4 & 67.0 & 50.7 & 47.3 \\
\midrule
ViT-S & Food-101 sym.\ 40\% & 39.8 & 53.5 & 40.5 & 38.1 \\
 & Food-101N web & 19.2 & 36.1 & 16.2 & 13.1 \\
ConvNeXt-T & Food-101 sym.\ 40\% & 39.8 & 52.5 & 31.4 & 31.9 \\
 & Food-101N web & 19.2 & 33.3 & 10.4 & 9.3 \\
\bottomrule
\end{tabular}
\end{table}

\subsection{Comparison with confident learning in every setting}
\label{app:cl}

Table~\ref{tab:cl} compares confident learning with the label-swap audit in every setting (Section~\ref{sec:cl}).

On CIFAR-10 the two methods separate wrong from right labels about equally well. The AUC of confident learning is
0.967--0.998 and that of the audit 0.958--0.995. Confident learning is ahead by 0.02--0.03 on the two milder human
label sets, the two are level on CIFAR-10N worst and 20\% symmetric noise, and the audit is ahead at 60\% symmetric
noise. Confident learning reads 70--101\% of the true rate and the audit 54--96\%, with flag precisions of 0.92--0.97
and 0.93--1.00. On clean CIFAR-10 they flag 0.3\% and 0.2\% of the examples. On CIFAR-100N confident learning is ahead
by 0.013 in AUC (0.925 against 0.912) and reads 34.7\% against 19.2\% at a true 41.3\%, with a precision of 0.84
against 0.96. On clean CIFAR-100 it flags 4.6\% of the examples and the audit 1.0\%. With symmetric noise on CIFAR-100
the two are level at 20\% (0.994 and 0.993) and the audit is ahead at 60\% (0.993 against 0.967). At 20\% the audit
also reads closer to the truth (19.2\% against 22.9\% at a true 19.7\%), because confident learning flags right labels
as well (precision 0.76). On CIFAR-10H confident learning keeps more of the separation (0.981 against 0.900) and reads
4.1\% against 1.4\% at a true 5.2\%. On Food-101 and Food-101N the AUC is 0.891--0.975 for confident learning and
0.876--0.984 for the audit, with flag precisions of 0.46--0.80 and 0.74--0.95.

Confident learning is given the training recipe, four further trainings and the predicted probabilities of every
training example, while the audit reads the 5{,}000 audited examples of the final model. Its reading comes from models
other than the one being audited, and it needs the training recipe and the means to train again, which a handed-over
model does not bring.

\begin{table*}[!tb]
\centering
\caption{Confident learning (CL, four further trainings on the whole training set for out-of-sample probabilities) against the label-swap audit
(LS) of the final model, ViT-S, on the same 5{,}000 training examples, mean of five runs. Reading: share of examples
flagged (\%). AUC of the CL self-confidence and of the label-swap score.}
\label{tab:cl}
\small\setlength{\tabcolsep}{5pt}
\begin{tabular}{lccccccc}
\toprule
 & & \multicolumn{2}{c}{Reading (\%)} & \multicolumn{2}{c}{Precision} & \multicolumn{2}{c}{AUC} \\
\cmidrule(lr){3-4}\cmidrule(lr){5-6}\cmidrule(lr){7-8}
Labels & Truth & CL & LS & CL & LS & CL & LS \\
\midrule
C10 clean & 0.0 & 0.3 & 0.2 & -- & -- & -- & -- \\
C10N aggregate & 8.3 & 5.8 & 4.5 & 0.917 & 0.934 & 0.989 & 0.958 \\
C10N random-1 & 16.5 & 14.4 & 11.8 & 0.943 & 0.956 & 0.991 & 0.972 \\
C10 sym.\ 20\% & 20.3 & 20.5 & 19.4 & 0.961 & 0.985 & 0.998 & 0.995 \\
C10N worst & 39.4 & 34.3 & 28.9 & 0.927 & 0.973 & 0.967 & 0.965 \\
C10 sym.\ 60\% & 58.4 & 54.9 & 54.8 & 0.973 & 0.996 & 0.984 & 0.995 \\
C100 clean & 0.0 & 4.6 & 1.0 & -- & -- & -- & -- \\
C100 sym.\ 20\% & 19.7 & 22.9 & 19.2 & 0.762 & 0.971 & 0.994 & 0.993 \\
C100N & 41.3 & 34.7 & 19.2 & 0.838 & 0.962 & 0.925 & 0.912 \\
C100 sym.\ 60\% & 60.4 & 58.7 & 56.2 & 0.898 & 0.995 & 0.967 & 0.993 \\
CIFAR-10H clean & 0.0 & 0.5 & 0.2 & -- & -- & -- & -- \\
CIFAR-10H random-1 & 5.2 & 4.1 & 1.4 & 0.831 & 0.893 & 0.981 & 0.900 \\
Food-101 sym.\ 40\% & 39.8 & 43.4 & 38.1 & 0.796 & 0.949 & 0.975 & 0.984 \\
Food-101N web & 19.2 & 19.8 & 13.1 & 0.461 & 0.740 & 0.891 & 0.876 \\
\midrule
Mean of the 11 noisy settings & & & & 0.846 & 0.941 & 0.969 & 0.958 \\
\bottomrule
\end{tabular}
\end{table*}

\subsection{Retraining: all settings}
\label{app:retrain}

Table~\ref{tab:retrain} gives the retraining results of Section~\ref{sec:retrain} for every setting and both backbones.
The gains are largest where the noise is heaviest: 28 to 47 points with 60\% symmetric noise, 11.6 and 17.0 points on
CIFAR-10N worst and 14.1 and 17.1 points on Food-101. The flagged lists beat the random lists in all 22 settings
($p = 4.8 \times 10^{-7}$). The removed examples carry wrong labels in 90 to 99.7\% of cases on CIFAR and Food-101 and
73 to 77\% on Food-101N.

\begin{table*}[t]
\centering
\caption{Retraining without the flagged examples, mean of five runs. Test accuracy (\%) of the original run (CE), of its
retraining without the examples flagged by the audit of its final model (flagged) and without as many random examples
(random). Removed: share of the training set removed; wrong: share of wrong labels among the removed examples (truth
used for scoring only). Bold: the highest of the three accuracies on noisy labels.}
\label{tab:retrain}
\small\setlength{\tabcolsep}{3pt}
\begin{tabular}{llccccc}
\toprule
Model & Labels & Removed & Wrong & CE & Flagged & Random \\
\midrule
ViT-S & C10N aggregate & 5.2 & 93.3 & 94.6 & \textbf{96.3} & 94.7 \\
 & C10N random-1 & 12.3 & 95.9 & 91.0 & \textbf{95.6} & 91.2 \\
 & C10N worst & 29.8 & 97.5 & 72.1 & \textbf{89.1} & 73.0 \\
 & C10 sym.\ 20\% & 19.3 & 98.2 & 90.6 & \textbf{98.1} & 90.7 \\
 & C10 sym.\ 60\% & 56.4 & 99.5 & 48.1 & \textbf{94.7} & 49.9 \\
 & C100N & 20.6 & 95.5 & 68.9 & \textbf{75.7} & 68.9 \\
 & C100 sym.\ 20\% & 19.9 & 95.8 & 84.8 & \textbf{90.2} & 84.5 \\
 & C100 sym.\ 60\% & 57.3 & 99.2 & 54.1 & \textbf{86.9} & 54.8 \\
 & CIFAR-10H random-1 & 1.4 & 90.7 & 96.6 & \textbf{97.0} & 96.7 \\
 & Food-101 sym.\ 40\% & 40.2 & 92.7 & 69.1 & \textbf{86.3} & 68.0 \\
 & Food-101N web & 14.2 & 72.6 & 80.3 & \textbf{82.2} & 79.9 \\
\midrule
ConvNeXt-T & C10N aggregate & 4.0 & 95.9 & 95.1 & \textbf{96.1} & 95.0 \\
 & C10N random-1 & 9.9 & 98.1 & 92.0 & \textbf{94.9} & 91.9 \\
 & C10N worst & 25.2 & 98.3 & 74.8 & \textbf{86.4} & 75.2 \\
 & C10 sym.\ 20\% & 18.5 & 99.2 & 92.4 & \textbf{97.9} & 92.5 \\
 & C10 sym.\ 60\% & 51.3 & 99.3 & 52.4 & \textbf{91.0} & 53.9 \\
 & C100N & 14.3 & 94.2 & 67.5 & \textbf{72.0} & 67.5 \\
 & C100 sym.\ 20\% & 18.9 & 97.8 & 85.1 & \textbf{90.0} & 84.3 \\
 & C100 sym.\ 60\% & 51.2 & 99.7 & 55.1 & \textbf{82.8} & 54.4 \\
 & CIFAR-10H random-1 & 1.0 & 90.4 & 96.4 & \textbf{96.7} & 96.4 \\
 & Food-101 sym.\ 40\% & 34.8 & 95.6 & 70.3 & \textbf{84.4} & 68.6 \\
 & Food-101N web & 10.1 & 76.6 & 80.6 & \textbf{82.0} & 80.1 \\
\midrule
\multicolumn{7}{l}{\emph{Clean labels}} \\
ViT-S & C10 clean & 0.2 & -- & 98.4 & 98.3 & 98.4 \\
 & C100 clean & 1.2 & -- & 90.8 & 90.7 & 90.9 \\
 & CIFAR-10H clean & 0.2 & -- & 98.1 & 98.0 & 98.1 \\
ConvNeXt-T & C10 clean & 0.0 & -- & 98.6 & 98.6 & 98.6 \\
 & C100 clean & 1.1 & -- & 90.8 & 90.9 & 90.9 \\
 & CIFAR-10H clean & 0.1 & -- & 97.9 & 97.9 & 97.9 \\
\bottomrule
\end{tabular}
\end{table*}

\subsection{Per-run results}
\label{app:perrun}

Tables~\ref{tab:perrun-auc} to~\ref{tab:perrun-retrain} give the values of every run behind the means reported in the
paper: the label-swap score at the last kept model on the 5{,}000 audited rows (AUC, reading, precision and recall;
Tables~\ref{tab:matrix}, \ref{tab:text} and~\ref{tab:cl}) and the test accuracy after retraining
(Table~\ref{tab:retrain}). The audited rows and their share of wrong labels (truth) are the same in all runs of a setting.

\begin{table}[h]
\centering
\caption{AUC of the label-swap score at the last kept model, per run (5{,}000 audited training examples).}
\label{tab:perrun-auc}
\small\setlength{\tabcolsep}{3pt}
\begin{tabular}{llcccccc}
\toprule
Model & Labels & Run 1 & Run 2 & Run 3 & Run 4 & Run 5 & Mean \\
\midrule
ViT-S & C10N aggregate & 0.958 & 0.952 & 0.961 & 0.959 & 0.961 & 0.958 \\
 & C10N random-1 & 0.973 & 0.968 & 0.972 & 0.975 & 0.973 & 0.972 \\
 & C10N worst & 0.967 & 0.963 & 0.965 & 0.969 & 0.961 & 0.965 \\
 & C10 sym.\ 20\% & 0.995 & 0.995 & 0.995 & 0.996 & 0.996 & 0.995 \\
 & C10 sym.\ 60\% & 0.995 & 0.994 & 0.995 & 0.996 & 0.995 & 0.995 \\
 & C100N & 0.913 & 0.910 & 0.908 & 0.912 & 0.915 & 0.912 \\
 & C100 sym.\ 20\% & 0.993 & 0.995 & 0.993 & 0.993 & 0.992 & 0.993 \\
 & C100 sym.\ 60\% & 0.993 & 0.992 & 0.993 & 0.993 & 0.992 & 0.993 \\
 & CIFAR-10H random-1 & 0.904 & 0.888 & 0.910 & 0.903 & 0.897 & 0.900 \\
 & Food-101 sym.\ 40\% & 0.983 & 0.985 & 0.985 & 0.984 & 0.984 & 0.984 \\
 & Food-101N web & 0.875 & 0.881 & 0.878 & 0.872 & 0.873 & 0.876 \\
ConvNeXt-T & C10N aggregate & 0.930 & 0.934 & 0.924 & 0.935 & 0.933 & 0.931 \\
 & C10N random-1 & 0.956 & 0.957 & 0.951 & 0.953 & 0.959 & 0.955 \\
 & C10N worst & 0.952 & 0.956 & 0.951 & 0.954 & 0.956 & 0.954 \\
 & C10 sym.\ 20\% & 0.992 & 0.992 & 0.993 & 0.993 & 0.993 & 0.993 \\
 & C10 sym.\ 60\% & 0.983 & 0.986 & 0.989 & 0.987 & 0.989 & 0.987 \\
 & C100N & 0.873 & 0.869 & 0.875 & 0.875 & 0.878 & 0.874 \\
 & C100 sym.\ 20\% & 0.990 & 0.992 & 0.992 & 0.991 & 0.991 & 0.991 \\
 & C100 sym.\ 60\% & 0.986 & 0.985 & 0.988 & 0.988 & 0.985 & 0.986 \\
 & CIFAR-10H random-1 & 0.856 & 0.838 & 0.842 & 0.828 & 0.829 & 0.839 \\
 & Food-101 sym.\ 40\% & 0.972 & 0.975 & 0.971 & 0.970 & 0.974 & 0.972 \\
 & Food-101N web & 0.860 & 0.853 & 0.859 & 0.862 & 0.868 & 0.860 \\
\midrule
RoBERTa & AG & 0.771 & 0.773 & 0.781 & 0.731 & 0.772 & 0.765 \\
 & SST-2 & 0.685 & 0.676 & 0.673 & 0.643 & 0.626 & 0.661 \\
 & QQP & 0.617 & 0.713 & 0.837 & 0.601 & 0.802 & 0.714 \\
 & MNLI & 0.720 & 0.700 & 0.740 & 0.711 & 0.692 & 0.713 \\
DeBERTa & AG & 0.710 & 0.737 & 0.709 & 0.703 & 0.721 & 0.716 \\
 & SST-2 & 0.782 & 0.735 & 0.628 & 0.662 & 0.710 & 0.703 \\
 & QQP & 0.647 & 0.708 & 0.584 & 0.668 & 0.617 & 0.645 \\
 & MNLI & 0.820 & 0.808 & 0.793 & 0.833 & 0.827 & 0.816 \\
Qwen3-1.7B & AG & 0.743 & 0.746 & 0.780 & 0.767 & 0.841 & 0.775 \\
\bottomrule
\end{tabular}
\end{table}

\begin{table}[h]
\centering
\caption{Reading of the noise rate (\% of the audited examples flagged) at the last kept model, per run, next to the truth.}
\label{tab:perrun-reading}
\small\setlength{\tabcolsep}{3pt}
\begin{tabular}{llcccccc}
\toprule
Model & Labels & Truth & Run 1 & Run 2 & Run 3 & Run 4 & Run 5 \\
\midrule
ViT-S & C10N aggregate & 8.3 & 4.7 & 4.7 & 4.4 & 4.4 & 4.3 \\
 & C10N random-1 & 16.5 & 12.0 & 11.5 & 11.7 & 12.3 & 11.7 \\
 & C10N worst & 39.4 & 29.3 & 28.7 & 29.0 & 29.3 & 28.5 \\
 & C10 sym.\ 20\% & 20.3 & 19.4 & 19.4 & 19.5 & 19.5 & 19.4 \\
 & C10 sym.\ 60\% & 58.4 & 54.9 & 55.0 & 55.0 & 55.0 & 54.4 \\
 & C100N & 41.3 & 19.1 & 19.1 & 19.4 & 19.5 & 18.9 \\
 & C100 sym.\ 20\% & 19.7 & 19.2 & 19.2 & 19.2 & 19.3 & 19.3 \\
 & C100 sym.\ 60\% & 60.4 & 56.3 & 56.2 & 56.4 & 55.8 & 56.3 \\
 & CIFAR-10H random-1 & 5.2 & 1.4 & 1.4 & 1.2 & 1.6 & 1.3 \\
 & Food-101 sym.\ 40\% & 39.8 & 38.3 & 37.9 & 38.3 & 38.4 & 37.5 \\
 & Food-101N web & 19.2 & 13.3 & 12.9 & 13.2 & 13.3 & 12.9 \\
ConvNeXt-T & C10N aggregate & 8.3 & 3.5 & 3.6 & 3.5 & 3.7 & 3.7 \\
 & C10N random-1 & 16.5 & 9.3 & 9.5 & 9.3 & 9.3 & 9.8 \\
 & C10N worst & 39.4 & 24.6 & 24.6 & 23.5 & 24.4 & 25.2 \\
 & C10 sym.\ 20\% & 20.3 & 18.6 & 18.8 & 18.5 & 18.8 & 18.8 \\
 & C10 sym.\ 60\% & 58.4 & 49.2 & 50.2 & 50.5 & 50.0 & 50.3 \\
 & C100N & 41.3 & 13.5 & 13.8 & 13.6 & 13.3 & 13.7 \\
 & C100 sym.\ 20\% & 19.7 & 18.1 & 18.2 & 18.2 & 18.2 & 18.2 \\
 & C100 sym.\ 60\% & 60.4 & 47.3 & 47.0 & 46.7 & 48.0 & 47.4 \\
 & CIFAR-10H random-1 & 5.2 & 0.8 & 0.9 & 1.0 & 1.1 & 0.8 \\
 & Food-101 sym.\ 40\% & 39.8 & 31.9 & 32.2 & 31.3 & 31.8 & 32.2 \\
 & Food-101N web & 19.2 & 9.2 & 9.7 & 9.1 & 9.3 & 9.0 \\
\midrule
RoBERTa & AG & 37.4 & 0.9 & 0.6 & 0.8 & 0.9 & 1.0 \\
 & SST-2 & 39.4 & 12.6 & 10.8 & 9.9 & 8.1 & 7.9 \\
 & QQP & 38.8 & 2.3 & 46.7 & 39.2 & 3.6 & 33.8 \\
 & MNLI & 40.1 & 5.7 & 3.3 & 5.8 & 3.7 & 3.8 \\
DeBERTa & AG & 37.4 & 2.5 & 3.7 & 3.4 & 1.1 & 2.2 \\
 & SST-2 & 39.4 & 26.5 & 10.6 & 5.7 & 7.8 & 7.2 \\
 & QQP & 38.8 & 4.5 & 2.2 & 4.2 & 5.3 & 2.8 \\
 & MNLI & 40.1 & 12.4 & 13.0 & 7.9 & 13.4 & 14.2 \\
Qwen3-1.7B & AG & 37.4 & 6.5 & 5.9 & 5.3 & 8.1 & 5.5 \\
\bottomrule
\end{tabular}
\end{table}

\begin{table*}[t]
\centering
\caption{Precision and recall of the flagged examples at the last kept model, per run.}
\label{tab:perrun-pr}
\small\setlength{\tabcolsep}{3pt}
\begin{tabular}{llcccccccccc}
\toprule
& & \multicolumn{5}{c}{Precision} & \multicolumn{5}{c}{Recall} \\
\cmidrule(lr){3-7}\cmidrule(lr){8-12}
Model & Labels & 1 & 2 & 3 & 4 & 5 & 1 & 2 & 3 & 4 & 5 \\
\midrule
ViT-S & C10N aggregate & 0.932 & 0.903 & 0.955 & 0.941 & 0.940 & 0.523 & 0.513 & 0.506 & 0.499 & 0.484 \\
 & C10N random-1 & 0.950 & 0.960 & 0.952 & 0.954 & 0.962 & 0.695 & 0.673 & 0.676 & 0.711 & 0.684 \\
 & C10N worst & 0.973 & 0.970 & 0.971 & 0.975 & 0.975 & 0.725 & 0.706 & 0.715 & 0.726 & 0.705 \\
 & C10 sym.\ 20\% & 0.991 & 0.985 & 0.977 & 0.988 & 0.985 & 0.947 & 0.943 & 0.940 & 0.949 & 0.941 \\
 & C10 sym.\ 60\% & 0.996 & 0.994 & 0.997 & 0.995 & 0.997 & 0.935 & 0.935 & 0.938 & 0.935 & 0.928 \\
 & C100N & 0.971 & 0.962 & 0.957 & 0.958 & 0.962 & 0.448 & 0.444 & 0.449 & 0.452 & 0.440 \\
 & C100 sym.\ 20\% & 0.971 & 0.973 & 0.975 & 0.971 & 0.968 & 0.947 & 0.947 & 0.949 & 0.952 & 0.947 \\
 & C100 sym.\ 60\% & 0.995 & 0.995 & 0.996 & 0.995 & 0.994 & 0.927 & 0.926 & 0.929 & 0.920 & 0.927 \\
 & CIFAR-10H random-1 & 0.875 & 0.897 & 0.935 & 0.890 & 0.866 & 0.240 & 0.233 & 0.221 & 0.279 & 0.221 \\
 & Food-101 sym.\ 40\% & 0.945 & 0.948 & 0.949 & 0.946 & 0.957 & 0.908 & 0.901 & 0.912 & 0.913 & 0.901 \\
 & Food-101N web & 0.739 & 0.756 & 0.752 & 0.725 & 0.729 & 0.514 & 0.508 & 0.520 & 0.503 & 0.490 \\
ConvNeXt-T & C10N aggregate & 0.954 & 0.972 & 0.955 & 0.934 & 0.968 & 0.396 & 0.417 & 0.405 & 0.410 & 0.429 \\
 & C10N random-1 & 0.987 & 0.975 & 0.987 & 0.979 & 0.972 & 0.559 & 0.563 & 0.557 & 0.554 & 0.581 \\
 & C10N worst & 0.977 & 0.985 & 0.990 & 0.976 & 0.987 & 0.610 & 0.615 & 0.592 & 0.605 & 0.631 \\
 & C10 sym.\ 20\% & 0.992 & 0.993 & 0.997 & 0.995 & 0.991 & 0.909 & 0.923 & 0.912 & 0.925 & 0.921 \\
 & C10 sym.\ 60\% & 0.988 & 0.992 & 0.995 & 0.994 & 0.997 & 0.832 & 0.852 & 0.859 & 0.850 & 0.858 \\
 & C100N & 0.944 & 0.942 & 0.925 & 0.962 & 0.949 & 0.309 & 0.314 & 0.305 & 0.310 & 0.314 \\
 & C100 sym.\ 20\% & 0.985 & 0.990 & 0.989 & 0.985 & 0.991 & 0.907 & 0.915 & 0.915 & 0.910 & 0.919 \\
 & C100 sym.\ 60\% & 0.996 & 0.998 & 0.999 & 0.998 & 0.997 & 0.780 & 0.777 & 0.773 & 0.792 & 0.782 \\
 & CIFAR-10H random-1 & 0.921 & 0.867 & 0.880 & 0.839 & 0.951 & 0.134 & 0.149 & 0.168 & 0.179 & 0.149 \\
 & Food-101 sym.\ 40\% & 0.969 & 0.971 & 0.973 & 0.969 & 0.972 & 0.777 & 0.784 & 0.765 & 0.773 & 0.785 \\
 & Food-101N web & 0.791 & 0.791 & 0.748 & 0.801 & 0.808 & 0.380 & 0.400 & 0.356 & 0.390 & 0.379 \\
\midrule
RoBERTa & AG & 0.909 & 0.966 & 0.881 & 0.936 & 0.902 & 0.021 & 0.015 & 0.020 & 0.024 & 0.025 \\
 & SST-2 & 0.728 & 0.793 & 0.757 & 0.647 & 0.649 & 0.232 & 0.218 & 0.191 & 0.133 & 0.129 \\
 & QQP & 0.678 & 0.537 & 0.709 & 0.706 & 0.726 & 0.040 & 0.645 & 0.715 & 0.065 & 0.632 \\
 & MNLI & 0.958 & 0.952 & 0.955 & 0.973 & 0.952 & 0.135 & 0.079 & 0.139 & 0.089 & 0.089 \\
DeBERTa & AG & 0.516 & 0.893 & 0.912 & 0.698 & 0.855 & 0.035 & 0.089 & 0.083 & 0.020 & 0.050 \\
 & SST-2 & 0.805 & 0.857 & 0.859 & 0.768 & 0.867 & 0.542 & 0.230 & 0.124 & 0.153 & 0.158 \\
 & QQP & 0.677 & 0.838 & 0.789 & 0.653 & 0.872 & 0.079 & 0.048 & 0.085 & 0.089 & 0.063 \\
 & MNLI & 0.965 & 0.954 & 0.965 & 0.978 & 0.958 & 0.299 & 0.310 & 0.191 & 0.326 & 0.339 \\
Qwen3-1.7B & AG & 0.828 & 0.848 & 0.868 & 0.791 & 0.949 & 0.144 & 0.135 & 0.123 & 0.172 & 0.139 \\
\bottomrule
\end{tabular}
\end{table*}

\begin{table*}[t]
\centering
\caption{Test accuracy per run (Runs 1--5) of the original run on its labels (CE), after retraining without the flagged examples (flagged) and without as many random examples (random), at the fixed endpoint of the original run.}
\label{tab:perrun-retrain}
\scriptsize\setlength{\tabcolsep}{2pt}
\begin{tabular}{llccccccccccccccc}
\toprule
& & \multicolumn{5}{c}{CE} & \multicolumn{5}{c}{Flagged} & \multicolumn{5}{c}{Random} \\
\cmidrule(lr){3-7}\cmidrule(lr){8-12}\cmidrule(lr){13-17}
Model & Labels & 1 & 2 & 3 & 4 & 5 & 1 & 2 & 3 & 4 & 5 & 1 & 2 & 3 & 4 & 5 \\
\midrule
ViT-S & C10N aggregate & 94.6 & 94.5 & 94.7 & 94.8 & 94.6 & 96.4 & 96.3 & 96.3 & 96.3 & 96.0 & 94.9 & 94.8 & 94.2 & 94.8 & 94.5 \\
 & C10N random-1 & 90.9 & 91.1 & 91.1 & 91.2 & 90.9 & 95.8 & 95.6 & 95.4 & 95.9 & 95.3 & 91.4 & 90.7 & 91.2 & 91.3 & 91.3 \\
 & C10N worst & 71.7 & 72.3 & 72.4 & 72.2 & 72.1 & 89.2 & 88.5 & 89.1 & 89.8 & 89.0 & 73.0 & 72.5 & 73.7 & 73.2 & 72.7 \\
 & C10 sym.\ 20\% & 90.3 & 90.5 & 90.7 & 90.8 & 90.6 & 98.0 & 98.1 & 98.0 & 98.2 & 98.0 & 90.7 & 90.4 & 90.6 & 90.9 & 91.0 \\
 & C10 sym.\ 60\% & 47.9 & 48.2 & 47.7 & 48.8 & 48.0 & 94.6 & 94.9 & 94.9 & 95.1 & 94.1 & 50.1 & 50.5 & 48.7 & 52.0 & 48.0 \\
 & C100N & 68.5 & 69.2 & 68.8 & 68.9 & 69.1 & 75.5 & 75.6 & 75.7 & 75.9 & 75.9 & 69.1 & 68.9 & 68.6 & 68.7 & 69.2 \\
 & C100 sym.\ 20\% & 84.7 & 84.8 & 84.8 & 84.8 & 84.7 & 90.0 & 90.1 & 90.4 & 90.2 & 90.1 & 84.6 & 84.4 & 84.1 & 84.4 & 84.9 \\
 & C100 sym.\ 60\% & 54.3 & 54.1 & 54.4 & 53.2 & 54.8 & 87.0 & 86.9 & 86.7 & 86.8 & 87.0 & 54.3 & 55.5 & 55.2 & 54.4 & 54.6 \\
 & CIFAR-10H random-1 & 96.7 & 96.8 & 96.6 & 96.7 & 96.5 & 97.0 & 97.0 & 96.9 & 97.2 & 96.9 & 96.7 & 96.7 & 96.8 & 96.7 & 96.7 \\
 & Food-101 sym.\ 40\% & 69.2 & 68.9 & 69.4 & 69.0 & 69.2 & 86.3 & 86.2 & 86.2 & 86.5 & 86.2 & 67.5 & 68.4 & 68.0 & 67.5 & 68.3 \\
 & Food-101N web & 80.3 & 80.2 & 80.5 & 80.3 & 80.5 & 82.4 & 82.2 & 82.0 & 82.4 & 81.9 & 80.0 & 79.8 & 79.9 & 79.6 & 80.0 \\
\cmidrule(lr){2-17}
 & C10 clean & 98.5 & 98.4 & 98.5 & 98.4 & 98.2 & 98.3 & 98.4 & 98.3 & 98.3 & 98.4 & 98.2 & 98.4 & 98.4 & 98.4 & 98.4 \\
 & C100 clean & 90.8 & 90.9 & 90.7 & 90.8 & 90.8 & 90.8 & 90.5 & 90.6 & 90.6 & 90.7 & 91.0 & 90.8 & 91.0 & 90.8 & 90.6 \\
 & CIFAR-10H clean & 98.1 & 98.2 & 98.1 & 98.1 & 98.0 & 98.1 & 98.2 & 97.9 & 98.1 & 98.0 & 98.1 & 98.0 & 98.1 & 98.3 & 98.0 \\
\midrule
ConvNeXt-T & C10N aggregate & 94.8 & 95.2 & 94.9 & 95.3 & 95.2 & 96.1 & 96.1 & 95.8 & 96.1 & 96.1 & 95.0 & 95.2 & 94.7 & 95.3 & 95.1 \\
 & C10N random-1 & 92.0 & 92.0 & 91.6 & 91.9 & 92.2 & 94.8 & 94.8 & 94.6 & 94.8 & 95.2 & 91.8 & 91.8 & 91.8 & 92.1 & 92.0 \\
 & C10N worst & 74.0 & 75.4 & 74.7 & 74.4 & 75.6 & 86.3 & 86.5 & 85.7 & 86.3 & 87.3 & 75.1 & 75.6 & 75.2 & 75.0 & 75.4 \\
 & C10 sym.\ 20\% & 92.4 & 92.5 & 92.4 & 92.5 & 92.3 & 97.8 & 97.9 & 98.0 & 97.9 & 97.8 & 92.3 & 92.6 & 92.7 & 92.7 & 92.4 \\
 & C10 sym.\ 60\% & 52.0 & 52.6 & 52.7 & 51.9 & 52.8 & 90.2 & 90.8 & 91.5 & 90.9 & 91.7 & 52.8 & 54.4 & 54.6 & 54.5 & 53.2 \\
 & C100N & 68.1 & 67.4 & 67.6 & 67.5 & 66.9 & 72.2 & 72.1 & 72.0 & 71.9 & 72.0 & 67.4 & 67.2 & 68.3 & 67.6 & 67.0 \\
 & C100 sym.\ 20\% & 84.8 & 85.1 & 85.4 & 85.2 & 85.1 & 90.0 & 90.1 & 90.3 & 90.1 & 89.7 & 84.3 & 83.8 & 84.9 & 84.0 & 84.3 \\
 & C100 sym.\ 60\% & 55.5 & 55.2 & 55.1 & 54.7 & 54.9 & 82.8 & 82.8 & 82.6 & 83.1 & 82.9 & 54.3 & 54.4 & 54.4 & 54.1 & 54.6 \\
 & CIFAR-10H random-1 & 96.4 & 96.4 & 96.5 & 96.2 & 96.3 & 96.6 & 96.7 & 96.8 & 96.7 & 96.8 & 96.4 & 96.4 & 96.6 & 96.1 & 96.6 \\
 & Food-101 sym.\ 40\% & 70.5 & 70.6 & 69.9 & 70.4 & 70.2 & 84.5 & 84.6 & 83.9 & 84.6 & 84.4 & 68.5 & 69.2 & 68.1 & 68.7 & 68.6 \\
 & Food-101N web & 80.8 & 80.7 & 80.7 & 80.7 & 80.3 & 82.0 & 81.8 & 82.0 & 82.3 & 82.0 & 80.1 & 80.2 & 79.9 & 80.3 & 80.1 \\
\cmidrule(lr){2-17}
 & C10 clean & 98.6 & 98.8 & 98.5 & 98.5 & 98.5 & 98.6 & 98.5 & 98.6 & 98.7 & 98.6 & 98.5 & 98.5 & 98.5 & 98.7 & 98.6 \\
 & C100 clean & 90.7 & 90.9 & 90.8 & 91.1 & 90.7 & 90.9 & 90.9 & 91.3 & 90.6 & 90.7 & 91.3 & 90.8 & 90.8 & 91.0 & 90.8 \\
 & CIFAR-10H clean & 98.0 & 97.9 & 98.1 & 97.8 & 97.7 & 97.7 & 98.0 & 98.0 & 97.9 & 97.9 & 97.9 & 98.0 & 97.9 & 97.7 & 97.8 \\
\bottomrule
\end{tabular}
\end{table*}

\subsection{Further analyses of the noise-rate reading}
\label{app:further}

\paragraph{CIFAR-100.} Two effects separate CIFAR-100 from CIFAR-10. First, on clean labels a model trained for two
epochs has not yet learned every class, and the audit flags the examples it still gets wrong. It flags 9.0\% and 10.0\%
of the examples (ViT-S and ConvNeXt-T) while the model misclassifies 8.8\% and 9.1\% of them, and 65\% and 60\% of the
misclassified examples are flagged against 3.6\% and 5.0\% of the correctly classified ones. With more training this
share falls to 1\%. Second, 35\% of the wrong human labels in CIFAR-100N name a class of the same superclass as the true
class, for example one kind of tree for another. These are the hardest errors to recover. After 40 epochs the label-swap
rule recovers 28\% of them against 54\% of the errors across superclasses (ViT-S; ConvNeXt-T: 18\% against 38\%), and
the lower recall is what lowers the reading through Eq.~\eqref{eq:identity}. The low reading does not come from the
number of audited examples. On the whole training set, with 500 examples per class instead of about 50, the epoch-40
readings are 20.6\% and 14.3\% against 19.2\% and 13.6\% on 5{,}000 examples, and the recall inside the superclass rises
only to 34\% and 21\%. Early in training the audit reads both the overall and the per-class noise rate. On the whole
training set the epoch-2 readings are 39.8\% and 38.9\% against a true 40.2\%, and the per-class readings correlate with
the per-class truth at 0.93 and 0.84 (0.89 and 0.80 with about 50 examples per class).

Two further observations bear on the rule itself. First, with 58\% symmetric noise most examples of each class carry a
wrong label, so the class means themselves are shifted. Counting negative margins then reads 4\% at epoch 2, whereas
the label-swap rule reads 59\%, because it compares the two labels of an example directly. Second, the closest other
class $a_i$ is the true class for 77--93\% of the wrong labels on CIFAR-10N and for 39--65\% on CIFAR-100N, so the
audit also proposes corrections.

\paragraph{CIFAR-10H.} With the label of one random annotator per image, 5.2\% of the audited labels are wrong. After
two epochs the scores separate them well (for ViT-S an AUC of 0.97 for the label-swap score, the feature margin and
the neighbour vote), but the threshold-free readings lie above the truth (8.9--9.7\% at 5.2\%, and 27\% for the
disagreement), because the model still disagrees with 27\% of its training labels. After fitting, the order of
CIFAR-10N returns (Table~\ref{tab:c10h}). For ViT-S the label-swap score keeps an AUC of 0.900 against 0.830 for the
loss and 0.863 for the feature margin, and its flag recovers 24\% of the wrong labels at a precision of 0.89, while
the feature margin, the neighbour vote and the prediction recover at most 7\%. ConvNeXt-T keeps a smaller lead in
AUC (0.839 against 0.832 for the feature margin and 0.799 for the loss), and its flag recovers 16\% of the wrong labels
at a precision of 0.89, against at most 4\% for the other flags. The reading is therefore a lower bound here as well. Most wrong labels of CIFAR-10H sit on images that people also find hard: 62\% of them fall on
images where fewer than 90\% of the annotators chose the true class, against 10\% of the right labels. The final ViT-S
model recovers 11--18\% of the wrong labels on these images and 39\% on the others. With the clean labels the final
models flag 0.2\% (ViT-S) and 0.1\% (ConvNeXt-T) of the examples.

\begin{table}[t]
\centering
\caption{CIFAR-10H with the label of one random annotator (5.2\% wrong among the 5{,}000 audited examples), final
model (epoch 80), mean of five runs. AUC for wrong vs.\ right labels, and the share of the wrong labels flagged by
the threshold-free flag, with the precision of the flag in parentheses.}
\label{tab:c10h}
\small\setlength{\tabcolsep}{3.5pt}
\begin{tabular}{lcccc}
\toprule
 & \multicolumn{2}{c}{ViT-S} & \multicolumn{2}{c}{ConvNeXt-T} \\
\cmidrule(lr){2-3}\cmidrule(lr){4-5}
Score & AUC & Flagged & AUC & Flagged \\
\midrule
Loss & 0.830 & -- & 0.799 & -- \\
Prediction $\neq$ label & -- & 0.00 & -- & 0.00 \\
Feature margin & 0.863 & 0.03 (1.00) & 0.832 & 0.01 (0.80) \\
Neighbour vote & 0.562 & 0.07 (1.00) & 0.540 & 0.04 (1.00) \\
Label-swap (ours) & 0.900 & 0.24 (0.89) & 0.839 & 0.16 (0.89) \\
\bottomrule
\end{tabular}
\end{table}

\paragraph{Readings from a held-out set.} The cross-fitting also gives every training run a held-out set: the fold
that its fold model did not see, with the same noisy labels. Read on that fold, the share of examples whose prediction
disagrees with the label exceeds the noise rate of the training set by 0.6--3.9 points on CIFAR-10, and confident
learning applied to the fold alone reads between 5.2 points below and 0.4 points above it. On CIFAR-10, a
class-imbalanced resample of 2{,}000 examples changes these readings by at most 1.2 points, and a random set of 300 by
at most 2.6 points on average, with a standard deviation across runs of up to 3.2 points. When the held-out images
carry Gaussian noise at severity 3 of the CIFAR-10-C corruptions \citep{hendrycks2019benchmarking}, the disagreement
over-reads the training-set noise rate by 13--19 points on CIFAR-10 and confident learning by up to 10 points,
although the training labels are unchanged. Impulse noise gives 6--8 points for the disagreement and between 4 points
below and 3 points above for confident learning. On CIFAR-100 Gaussian noise moves the disagreement by 23 to 46 points
and confident learning by 5 to 29 points, and on CIFAR-10H by 20 to 22 and 7 to 10 points. The audit uses no held-out
set, so the way such
a set is drawn does not enter its reading.

\subsection{Cost}
\label{app:cost}

Table~\ref{tab:cost} lists the wall-clock time of the audit and of training on one GPU. The
audit takes one pass over the audited examples to form the class means and one further pass per label considered, so
it grows linearly with the number of examples. For 5{,}000 examples the label-swap audit takes 14 to 27 seconds, about
one hundredth of the time it took to train the model. The audit of 1{,}000 classes is timed in
Appendix~\ref{app:repro-abl}.

\begin{table}[t]
\centering
\caption{Wall-clock time on one RTX 4090, mean over all audited runs.}
\label{tab:cost}
\small\setlength{\tabcolsep}{4pt}
\begin{tabular}{lcc}
\toprule
 & ViT-S/16 & ConvNeXt-T \\
\midrule
Margins, 500 ex. & 0.7 s & 1.3 s \\
Margins, 5{,}000 ex. & 6.8 s & 13.2 s \\
Margins, 50{,}000 ex. & 68 s & 131 s \\
Label-swap, 5{,}000 ex. & 14 s & 26 s \\
Label-swap, 50{,}000 ex. & 146 s & 272 s \\
\midrule
Training, 40 epochs & 27--33 min & 51--68 min \\
\bottomrule
\end{tabular}
\end{table}

\FloatBarrier
\section{Extension experiments: results}
\label{app:ext-results}

\subsection{Other ways of choosing what to remove}
\label{app:cleaning}

For the 11 noisy ViT-S settings and each of the five runs, two further lists are removed before retraining with the
recipe and seed of the original run. \emph{Loss}: the examples with the highest cross-entropy loss under the final
model, as many as the audit flags in that run. \emph{Confident learning}: the label issues that confident learning
\citep{northcutt2021confident} finds with its default rule from out-of-sample predicted probabilities of the whole
training set, obtained from four further trainings on four folds (10 epochs each, the original recipe otherwise), as
many as it finds. Table~\ref{tab:cleaning} reports the test accuracy at the fixed endpoint of the original run. Over
the 11 settings (Wilcoxon signed-rank test, unit = setting), removing the audit's flags is above removing by loss in
10 ($p = 2.0 \times 10^{-3}$) and level with confident learning ($p = 0.70$); all three removals are above the original
run in all 11 settings ($p = 9.8 \times 10^{-4}$).

\begin{table*}[t]
\centering
\caption{Retraining after three ways of choosing what to remove, ViT-S, mean of five runs. Removed: share of the
training set removed by the audit (flagged; removal by loss removes as many) and by confident learning (CL). Wrong:
share of wrong labels among the removed examples (truth used for scoring only). Accuracy: test accuracy (\%) of the
original run (CE) and of the retraining without the flagged examples, the highest-loss examples and the CL label
issues. Bold: the highest accuracy.}
\label{tab:cleaning}
\vspace{3pt}
\small\setlength{\tabcolsep}{3pt}
\begin{tabular}{llccccccccc}
\toprule
 & & \multicolumn{2}{c}{Removed (\%)} & \multicolumn{3}{c}{Wrong (\%)} & \multicolumn{4}{c}{Test accuracy (\%)} \\
\cmidrule(lr){3-4}\cmidrule(lr){5-7}\cmidrule(lr){8-11}
Model & Labels & Flagged & CL & Flagged & Loss & CL & CE & Flagged & Loss & CL \\
\midrule
ViT-S & C10N aggregate & 5.2 & 6.3 & 93.3 & 72.6 & 91.4 & 94.6 & 96.3 & 96.0 & \textbf{97.0} \\
 & C10N random-1 & 12.3 & 15.1 & 95.9 & 69.1 & 93.5 & 91.0 & 95.6 & 93.7 & \textbf{96.8} \\
 & C10N worst & 29.8 & 35.0 & 97.5 & 69.4 & 93.1 & 72.1 & 89.1 & 81.0 & \textbf{90.5} \\
 & C10 sym.\ 20\% & 19.3 & 20.4 & 98.2 & 74.9 & 95.2 & 90.6 & 98.1 & 96.5 & \textbf{98.3} \\
 & C10 sym.\ 60\% & 56.4 & 56.4 & 99.5 & 76.4 & 97.1 & 48.1 & \textbf{94.7} & 75.5 & 94.0 \\
 & C100N & 20.6 & 33.6 & 95.5 & 81.8 & 84.2 & 68.9 & 75.7 & 73.0 & \textbf{78.9} \\
 & C100 sym.\ 20\% & 19.9 & 23.0 & 95.8 & 83.0 & 77.1 & 84.8 & \textbf{90.2} & 89.5 & 89.6 \\
 & C100 sym.\ 60\% & 57.3 & 58.2 & 99.2 & 84.6 & 89.7 & 54.1 & \textbf{86.9} & 79.8 & 83.2 \\
 & CIFAR-10H random-1 & 1.4 & 4.0 & 90.8 & 87.2 & 84.1 & 96.6 & 97.0 & 97.0 & \textbf{97.6} \\
 & Food-101 sym.\ 40\% & 40.2 & 43.3 & 92.7 & 73.0 & 79.9 & 69.1 & \textbf{86.3} & 80.8 & 84.8 \\
 & Food-101N web & 14.2 & 19.6 & 72.6 & 56.4 & 44.7 & 80.3 & 82.2 & 81.4 & \textbf{82.3} \\
\midrule
\multicolumn{2}{l}{Mean of 11 settings} & & & 93.7 & 75.3 & 84.6 & 77.3 & 90.2 & 85.8 & 90.3 \\
\bottomrule
\end{tabular}
\end{table*}

\subsection{Inversion attack on the message of the split audit}
\label{app:privacy}

\paragraph{What the owner of the data computes.} The split audit of Section~\ref{sec:method} asks the owner for
statistics computed from the monitored blocks. Cheaper inputs do not carry the signal once the labels are fitted. Over
the 11 noisy ViT-S settings, the same audit computed from the classifier head alone, that is from the final features and
the output probabilities, reaches an AUC of 0.957 after two epochs but 0.654 at the last kept model, where it reads
4\% of the true noise rate. The last block gives 0.932 and the four monitored blocks 0.958 at the last kept model, and the
four blocks are higher than both in all 11 settings (Wilcoxon signed-rank test, $p = 9.8 \times 10^{-4}$ each). This
agrees with the argument of Figure~\ref{fig:mechanism} that fitting rewrites the final features more than the
activations of the earlier blocks.

The messages, the attacker and the decoder are described in Appendix~\ref{app:privacy-setup}.

\paragraph{Scores.} The decoder is applied to the messages of the 5{,}000 audited training examples of the owner.
PSNR and SSIM (Gaussian window of 11 pixels, $\sigma = 1.5$) compare each reconstruction with its image. For
re-identification, the reconstruction of each of the first 1{,}000 examples is compared with the 1{,}000 images by
Euclidean distance in pixel space, and the attack succeeds when the true image is the nearest. Truth labels are not used.
Table~\ref{tab:privacy} gives the results.

\begin{table*}[t]
\centering
\caption{Inversion attack, ViT-S, final model, mean of five runs. Re-identification: share (\%) of examples whose
reconstruction is nearest to their own image among 1{,}000 candidates (chance 0.1\%). PSNR (dB) of the reconstruction.
Label: mean image of the observed class. Message: the $2K + 4$ numbers of the split audit. Sketch: 4{,}096-number
sketch of the allocation. Features: the 384 final features.}
\label{tab:privacy}
\vspace{3pt}
\small\setlength{\tabcolsep}{4pt}
\begin{tabular}{llcccccccc}
\toprule
 & & \multicolumn{4}{c}{Re-identification (\%)} & \multicolumn{4}{c}{PSNR (dB)} \\
\cmidrule(lr){3-6}\cmidrule(lr){7-10}
Model & Labels & Label & Message & Sketch & Features & Label & Message & Sketch & Features \\
\midrule
ViT-S & C10N aggregate & 0.00 & 0.34 & 3.72 & 4.56 & 13.0 & 13.0 & 13.6 & 13.6 \\
 & C10N random-1 & 0.10 & 0.20 & 3.32 & 4.36 & 12.9 & 12.9 & 13.4 & 13.3 \\
 & C10N worst & 0.10 & 0.28 & 3.18 & 3.48 & 12.8 & 12.6 & 13.5 & 13.2 \\
 & C10 sym.\ 20\% & 0.10 & 0.34 & 3.14 & 3.26 & 12.9 & 12.7 & 13.4 & 13.2 \\
 & C10 sym.\ 60\% & 0.20 & 0.26 & 2.56 & 2.88 & 12.6 & 12.2 & 13.3 & 13.0 \\
 & C100N & 0.50 & 1.26 & 4.10 & 3.16 & 12.6 & 12.3 & 13.2 & 12.8 \\
 & C100 sym.\ 20\% & 0.20 & 2.26 & 3.74 & 3.36 & 12.6 & 12.4 & 13.2 & 12.7 \\
 & C100 sym.\ 60\% & 0.20 & 0.88 & 3.14 & 1.30 & 12.1 & 11.9 & 13.0 & 12.1 \\
 & CIFAR-10H random-1 & 0.10 & 0.32 & 3.78 & 4.56 & 13.0 & 13.1 & 13.6 & 13.7 \\
 & Food-101 sym.\ 40\% & 0.00 & 0.98 & 1.90 & 0.98 & 12.2 & 11.4 & 12.2 & 11.5 \\
 & Food-101N web & 0.10 & 0.30 & 1.60 & 1.22 & 12.3 & 11.5 & 12.0 & 11.8 \\
\midrule
\multicolumn{2}{l}{Mean of 11 settings} & 0.15 & 0.67 & 3.11 & 3.01 & 12.6 & 12.4 & 13.1 & 12.8 \\
\bottomrule
\end{tabular}
\end{table*}

\subsection{Text classification}
\label{app:text}

Table~\ref{tab:text} lists the AUC of every score on the final model. The label-swap score is the
highest final-model score in all nine settings, with a mean of 0.723 against 0.53 to 0.58 for the others (Wilcoxon
signed-rank test over the settings, $p = 0.0039$ against each, the smallest value nine settings allow). TracIn over the
kept checkpoints reaches 0.760 on average and does not differ significantly ($p = 0.12$); on QQP the audit is higher.
After one epoch, before the wrong labels are fitted, the reading on the GLUE tasks is 40.2 to 46.8\% at true rates of
38.8 to 40.1\%. Once they are fitted it falls to 1 to 12\% (25\% for RoBERTa on QQP, where three of five
runs do not fit the wrong labels), with precisions of 0.67 to 0.96, so on text the reading of a fitted model is a lower
bound. An audit of 5{,}000 examples takes 30 seconds for RoBERTa, 44 seconds for DeBERTa and 16 minutes for Qwen3-1.7B,
against 53 to 85 and 129 minutes of training.

\begin{table*}[t]
\centering
\caption{AUC for wrong vs.\ right labels at the final model on text, 5{,}000 training examples, mean of five runs.
LS: label-swap (ours); FM: feature margin; NV: neighbour vote (kNN-10); GN: gradient norm; SI: self-influence; TracIn
sums over the kept checkpoints. AG: NoisyAG-News worst human labels; the GLUE tasks carry 40\% symmetric noise.}
\label{tab:text}
\small\setlength{\tabcolsep}{3pt}
\begin{tabular}{llccccccc}
\toprule
 & & \multicolumn{6}{c}{Final model only} & History \\
\cmidrule(lr){3-8}\cmidrule(lr){9-9}
Model & Data & LS & Loss & FM & NV & GN & SI & TracIn \\
\midrule
RoBERTa & AG & \textbf{0.765} & 0.550 & 0.576 & 0.504 & 0.541 & 0.538 & 0.845 \\
 & SST-2 & \textbf{0.661} & 0.566 & 0.571 & 0.560 & 0.565 & 0.564 & 0.679 \\
 & QQP & \textbf{0.714} & 0.655 & 0.651 & 0.594 & 0.597 & 0.596 & 0.683 \\
 & MNLI & \textbf{0.713} & 0.572 & 0.582 & 0.502 & 0.568 & 0.565 & 0.838 \\
DeBERTa & AG & \textbf{0.716} & 0.460 & 0.491 & 0.505 & 0.476 & 0.474 & 0.841 \\
 & SST-2 & \textbf{0.703} & 0.571 & 0.569 & 0.562 & 0.574 & 0.578 & 0.706 \\
 & QQP & \textbf{0.645} & 0.513 & 0.473 & 0.500 & 0.504 & 0.502 & 0.591 \\
 & MNLI & \textbf{0.816} & 0.536 & 0.565 & 0.523 & 0.538 & 0.537 & 0.818 \\
Qwen3-1.7B & AG & \textbf{0.775} & 0.654 & 0.727 & 0.559 & 0.652 & 0.653 & 0.838 \\
\midrule
\multicolumn{2}{l}{Mean} & \textbf{0.723} & 0.564 & 0.578 & 0.534 & 0.557 & 0.556 & 0.760 \\
\bottomrule
\end{tabular}
\end{table*}

\subsection{Asymmetric noise}
\label{app:asym}

Table~\ref{tab:asym} gives the scores of the asymmetric-noise settings of Section~\ref{sec:cost}. In each affected class a
fixed 40\% of the training labels, drawn with a fixed seed, is moved to the partner class. On CIFAR-10 the affected
classes are truck, bird, deer, cat and dog (truck to automobile, bird to airplane, deer to horse, cat to dog and dog to
cat). On CIFAR-100 every class is affected and moves to the next class, in label order, of its superclass. After 40
epochs the model predicts 99.3--100\% of the wrong labels on CIFAR-10 and all of them on CIFAR-100 (on 2{,}000 wrongly
labelled training examples tracked during training), so the wrong labels are fitted. The audit, the
other scores and confident learning are computed exactly as for the other settings, on the same 5{,}000 training
examples of each run.

\begin{table}[h]
\centering
\caption{Asymmetric noise (40\% of each affected class), ViT-S, last kept model (epoch 40), same 5{,}000 training
examples, mean and range over five runs. Upper block: AUC for wrong vs.\ right labels. Lower block: flags of the
label-swap audit (LS) and of confident learning (CL), reading in \% of the audited examples.}
\label{tab:asym}
\small\setlength{\tabcolsep}{4pt}
\begin{tabular}{lcc}
\toprule
 & CIFAR-10 (20.1\% wrong) & CIFAR-100 (39.9\% wrong) \\
\midrule
\multicolumn{3}{l}{\emph{AUC, final model only}} \\
Label-swap (ours) & \textbf{0.992} (0.988--0.994) & \textbf{0.935} (0.928--0.942) \\
Loss & 0.809 (0.803--0.814) & 0.590 (0.575--0.601) \\
Feature margin & 0.956 (0.951--0.965) & 0.796 (0.790--0.802) \\
Neighbour vote & 0.584 (0.560--0.626) & 0.629 (0.623--0.632) \\
Gradient norm & 0.823 (0.819--0.826) & 0.584 (0.569--0.595) \\
Self-influence & 0.822 (0.818--0.826) & 0.584 (0.569--0.595) \\
\multicolumn{3}{l}{\emph{AUC, further inputs}} \\
TracIn, 5 checkpoints & 0.909 (0.883--0.925) & 0.709 (0.689--0.733) \\
Confident learning & 0.919 (0.909--0.927) & 0.691 (0.682--0.698) \\
\midrule
LS reading (\%) & 17.9 (16.8--18.7) & 26.7 (26.0--27.8) \\
CL reading (\%) & 17.4 (17.1--17.7) & 47.4 (45.4--49.0) \\
LS precision & 0.960 (0.944--0.971) & 0.952 (0.947--0.955) \\
CL precision & 0.772 (0.756--0.780) & 0.561 (0.547--0.574) \\
LS recall & 0.855 (0.787--0.892) & 0.638 (0.622--0.666) \\
CL recall & 0.666 (0.658--0.685) & 0.667 (0.654--0.675) \\
\bottomrule
\end{tabular}
\end{table}

Table~\ref{tab:asym-perrun} gives the label-swap audit of each run.

\begin{table}[h]
\centering
\caption{Asymmetric noise, per run: the label-swap audit at the last kept model (ViT-S, 5{,}000 audited examples).}
\label{tab:asym-perrun}
\small\setlength{\tabcolsep}{4pt}
\begin{tabular}{llcccccc}
\toprule
Labels & & Run 1 & Run 2 & Run 3 & Run 4 & Run 5 & Mean \\
\midrule
C10 asym.\ 40\% & AUC & 0.988 & 0.993 & 0.994 & 0.994 & 0.993 & 0.992 \\
 & Reading (\%) & 16.8 & 17.8 & 18.5 & 18.7 & 17.9 & 17.9 \\
 & Precision & 0.944 & 0.970 & 0.971 & 0.956 & 0.962 & 0.960 \\
 & Recall & 0.787 & 0.856 & 0.892 & 0.888 & 0.855 & 0.855 \\
\midrule
C100 asym.\ 40\% & AUC & 0.942 & 0.940 & 0.928 & 0.932 & 0.930 & 0.935 \\
 & Reading (\%) & 27.8 & 26.7 & 26.4 & 26.7 & 26.0 & 26.7 \\
 & Precision & 0.955 & 0.955 & 0.951 & 0.947 & 0.954 & 0.952 \\
 & Recall & 0.666 & 0.639 & 0.629 & 0.634 & 0.622 & 0.638 \\
\bottomrule
\end{tabular}
\end{table}

\subsection{A backbone with batch normalisation}
\label{app:resnet}

Table~\ref{tab:resnet} gives every score for ResNet-50 on CIFAR-10N worst and CIFAR-100N (Section~\ref{sec:cost}), and
Table~\ref{tab:resnet-perrun} the label-swap audit of each run. The final models reach a test accuracy of 69.4\% and
63.7\%.

\begin{table}[h]
\centering
\caption{ResNet-50, a backbone with batch normalisation, at the final model (CIFAR-10N worst and CIFAR-100N, mean of
five runs, 5{,}000 audited examples). Top: AUC for wrong vs.\ right labels, columns as in Table~\ref{tab:matrix}.
Bottom: the flags of the label-swap test. Fitted: share of the wrong labels that the final model predicts. True: share
of wrong labels among the audited examples.}
\label{tab:resnet}
\small\setlength{\tabcolsep}{3.5pt}
\begin{tabular}{lccccccc}
\toprule
 & \multicolumn{6}{c}{Final model only} & History \\
\cmidrule(lr){2-7}\cmidrule(lr){8-8}
Labels & LS & Loss & FM & NV & GN & SI & TracIn \\
\midrule
C10N worst & \textbf{0.958} & 0.700 & 0.852 & 0.749 & 0.695 & 0.693 & 0.874 \\
C100N & \textbf{0.905} & 0.677 & 0.863 & 0.757 & 0.689 & 0.684 & 0.853 \\
\bottomrule
\end{tabular}

\vspace{6pt}
\begin{tabular}{lccccc}
\toprule
Labels & Fitted (\%) & True (\%) & Reading (\%) & Precision & Recall \\
\midrule
C10N worst & 97.7 & 39.4 & 30.8 & 0.948 & 0.743 \\
C100N & 99.9 & 41.3 & 19.8 & 0.942 & 0.450 \\
\bottomrule
\end{tabular}
\end{table}

\begin{table}[h]
\centering
\caption{ResNet-50, per run: the label-swap audit at the final model (5{,}000 audited examples).}
\label{tab:resnet-perrun}
\small\setlength{\tabcolsep}{4pt}
\begin{tabular}{llcccccc}
\toprule
Labels & & Run 1 & Run 2 & Run 3 & Run 4 & Run 5 & Mean \\
\midrule
C10N worst & AUC & 0.961 & 0.957 & 0.960 & 0.953 & 0.959 & 0.958 \\
 & Reading (\%) & 30.5 & 31.3 & 30.2 & 31.2 & 30.9 & 30.8 \\
 & Precision & 0.951 & 0.946 & 0.960 & 0.942 & 0.944 & 0.948 \\
 & Recall & 0.738 & 0.752 & 0.737 & 0.746 & 0.742 & 0.743 \\
\midrule
C100N & AUC & 0.905 & 0.907 & 0.905 & 0.906 & 0.904 & 0.905 \\
 & Reading (\%) & 20.1 & 19.7 & 19.8 & 19.6 & 19.7 & 19.8 \\
 & Precision & 0.942 & 0.940 & 0.943 & 0.945 & 0.938 & 0.942 \\
 & Recall & 0.457 & 0.448 & 0.451 & 0.448 & 0.446 & 0.450 \\
\bottomrule
\end{tabular}
\end{table}

\subsection{One thousand classes}
\label{app:inat}

Table~\ref{tab:inat} gives every score and the flags for the 1{,}000 species of iNaturalist 2021
(Section~\ref{sec:cost}), Table~\ref{tab:inat-epochs} the label-swap audit at every kept model and
Table~\ref{tab:inat-perrun} the audit of each run. Every audit covers the whole training set of 50{,}000 examples. The
final models reach a test accuracy of 59.0\% and 49.7\%. The audit from a random subset of the coordinates is given
with the other backbones in Tables~\ref{tab:coords} and~\ref{tab:coords-perrun}.

\begin{table}[h]
\centering
\caption{1{,}000 species of iNaturalist 2021, ViT-S, final model (epoch 40), whole training set (50{,}000 examples),
mean of five runs. Top: AUC for wrong vs.\ right labels, columns as in Table~\ref{tab:matrix}. Bottom: the flags of the
label-swap test. Fitted: share of the wrong labels that the final model predicts. True: share of wrong labels among the
audited examples.}
\label{tab:inat}
\small\setlength{\tabcolsep}{3.5pt}
\begin{tabular}{lccccccc}
\toprule
 & \multicolumn{6}{c}{Final model only} & History \\
\cmidrule(lr){2-7}\cmidrule(lr){8-8}
Labels & LS & Loss & FM & NV & GN & SI & TracIn \\
\midrule
Sym.\ 40\% & \textbf{0.991} & 0.763 & 0.966 & 0.971 & 0.737 & 0.733 & 0.888 \\
Asym.\ 40\% & \textbf{0.852} & 0.564 & 0.778 & 0.664 & 0.559 & 0.559 & 0.605 \\
\bottomrule
\end{tabular}

\vspace{6pt}
\begin{tabular}{lccccc}
\toprule
Labels & Fitted (\%) & True (\%) & Reading (\%) & Precision & Recall \\
\midrule
Sym.\ 40\% & 100.0 & 40.0 & 42.5 & 0.916 & 0.974 \\
Asym.\ 40\% & 100.0 & 40.0 & 23.4 & 0.837 & 0.488 \\
\bottomrule
\end{tabular}
\end{table}

\begin{table}[h]
\centering
\caption{1{,}000 species of iNaturalist 2021: the label-swap audit at every kept model (ViT-S, whole training set, all
coordinates, mean of five runs). Fitted: share of the wrong labels that the model predicts. True share of wrong labels
40.0\%.}
\label{tab:inat-epochs}
\small\setlength{\tabcolsep}{4pt}
\begin{tabular}{llccccc}
\toprule
Labels & & Epoch 2 & Epoch 5 & Epoch 10 & Epoch 20 & Epoch 40 \\
\midrule
Sym.\ 40\% & Fitted (\%) & 0.2 & 5.2 & 98.6 & 100.0 & 100.0 \\
 & AUC & 0.984 & 0.989 & 0.990 & 0.991 & 0.991 \\
 & Reading (\%) & 58.5 & 51.8 & 43.4 & 42.8 & 42.5 \\
 & Precision & 0.680 & 0.769 & 0.901 & 0.910 & 0.916 \\
 & Recall & 0.995 & 0.996 & 0.978 & 0.975 & 0.974 \\
\midrule
Asym.\ 40\% & Fitted (\%) & 22.6 & 59.6 & 95.1 & 99.6 & 100.0 \\
 & AUC & 0.756 & 0.835 & 0.854 & 0.839 & 0.852 \\
 & Reading (\%) & 55.3 & 43.9 & 31.1 & 25.9 & 23.4 \\
 & Precision & 0.564 & 0.680 & 0.786 & 0.802 & 0.837 \\
 & Recall & 0.781 & 0.745 & 0.610 & 0.518 & 0.488 \\
\bottomrule
\end{tabular}
\end{table}

\begin{table}[h]
\centering
\caption{1{,}000 species of iNaturalist 2021, per run: the label-swap audit at the final model (ViT-S, whole training
set, all coordinates).}
\label{tab:inat-perrun}
\small\setlength{\tabcolsep}{4pt}
\begin{tabular}{llcccccc}
\toprule
Labels & & Run 1 & Run 2 & Run 3 & Run 4 & Run 5 & Mean \\
\midrule
Sym.\ 40\% & AUC & 0.991 & 0.991 & 0.991 & 0.991 & 0.991 & 0.991 \\
 & Reading (\%) & 42.5 & 42.4 & 42.6 & 42.5 & 42.5 & 42.5 \\
 & Precision & 0.917 & 0.917 & 0.913 & 0.916 & 0.918 & 0.916 \\
 & Recall & 0.973 & 0.973 & 0.973 & 0.974 & 0.976 & 0.974 \\
\midrule
Asym.\ 40\% & AUC & 0.853 & 0.852 & 0.850 & 0.852 & 0.852 & 0.852 \\
 & Reading (\%) & 23.2 & 23.6 & 23.3 & 23.3 & 23.4 & 23.4 \\
 & Precision & 0.838 & 0.837 & 0.835 & 0.837 & 0.835 & 0.837 \\
 & Recall & 0.485 & 0.494 & 0.486 & 0.488 & 0.489 & 0.488 \\
\bottomrule
\end{tabular}
\end{table}

\FloatBarrier
\section{Ablations: results}
\label{app:ablation}

Table~\ref{tab:ablation} collects the ablations of Sections~\ref{sec:private} and~\ref{sec:cost}. Each row asks one
question about the audit and answers it with one measurement: the AUC for wrong vs.\ right labels at the final model of
CIFAR-10N worst, where 99\% of the wrong labels are fitted. The audit as it is used in the paper reaches 0.965 there.
Tables~\ref{tab:abl-layers}, \ref{tab:abl-precbatch} and~\ref{tab:abl-samples} give the full numbers for the monitored
blocks, the training precision and batch size, and the sample size, on both backbones and on CIFAR-100N where they were
run. Tables~\ref{tab:coords} and~\ref{tab:coords-perrun} give the audit from a random subset of the coordinates.

\begin{table}[h]
\centering
\caption{Ablations of the label-swap audit, one question per row. AUC for wrong vs.\ right labels at the final model
(ViT-S, CIFAR-10N worst, mean of five runs, 5{,}000 audited examples). The row on 1{,}000 classes uses its own data
(Appendix~\ref{app:inat}). The last row uses the allocation margin of Eq.~\eqref{eq:margin} on random subsets of the
training set.}
\label{tab:ablation}
\small\setlength{\tabcolsep}{4pt}
\begin{tabular}{p{0.27\textwidth}p{0.32\textwidth}l}
\toprule
Question & What is changed & AUC \\
\midrule
\raggedright Reference & \raggedright the audit as used in the paper & 0.965 \\
\midrule
\raggedright Is the normalisation needed? & \raggedright raw magnitudes $|g_i|$, not normalised & 0.784 \\
\raggedright Are the absolute values needed? & \raggedright signed gradient, normalised & 0.765 \\
\raggedright Does the classifier head suffice? & \raggedright classifier head only & 0.618 \\
\raggedright How many blocks to monitor? & \raggedright the last 1 / 2 / 4 / 8 / 12 blocks & 0.919 / 0.947 / 0.965 / 0.968 / 0.966 \\
\raggedright What does the second pass add? & \raggedright margin of the observed label alone & 0.960 \\
\raggedright Does the training precision matter? & \raggedright FP32 in place of bfloat16 & 0.964 \\
\raggedright Does the batch size matter? & \raggedright batch size 64 / 256 in place of 128 & 0.965 / 0.964 \\
\raggedright Are all coordinates needed? & \raggedright a random 10\% / 1\% / 0.1\% of the coordinates & 0.965 / 0.960 / 0.918 \\
\raggedright Does batch normalisation in the backbone matter? & \raggedright ResNet-50 in place of ViT-S & 0.958 \\
\raggedright Does the audit carry over to 1{,}000 classes? & \raggedright 1{,}000 iNaturalist species, symmetric / asymmetric 40\%, all 50{,}000 examples & 0.991 / 0.852 \\
\raggedright How many examples to audit? & \raggedright 500 / 5{,}000 / all 50{,}000 examples & 0.963 / 0.961 / 0.963 \\
\bottomrule
\end{tabular}
\end{table}

\begin{table}[h]
\centering
\caption{Monitored blocks: the label-swap audit with the last $k$ blocks monitored, after epoch 2 and at the final model (epoch 40), mean of five runs, 5{,}000 audited examples. Reading in \% (true share 39.4\% on CIFAR-10N worst and 41.3\% on CIFAR-100N). All other results of the paper use $k = 4$.}
\label{tab:abl-layers}
\small\setlength{\tabcolsep}{4pt}
\begin{tabular}{llccccccc}
\toprule
 & & & \multicolumn{2}{c}{Epoch 2} & \multicolumn{4}{c}{Epoch 40} \\
\cmidrule(lr){4-5}\cmidrule(lr){6-9}
Model & Labels & $k$ & AUC & Reading & AUC & Reading & Precision & Recall \\
\midrule
ViT-S & C10N worst & 1 & 0.990 & 38.2 & 0.919 & 17.8 & 0.981 & 0.444 \\
 &  & 2 & 0.992 & 38.7 & 0.947 & 24.0 & 0.985 & 0.600 \\
 &  & 4 & 0.992 & 38.8 & 0.965 & 28.9 & 0.973 & 0.716 \\
 &  & 8 & 0.991 & 38.9 & 0.968 & 31.7 & 0.955 & 0.769 \\
 &  & 12 & 0.990 & 39.0 & 0.966 & 32.3 & 0.950 & 0.779 \\
\midrule
ViT-S & C100N & 1 & 0.946 & 39.5 & 0.882 & 10.2 & 0.975 & 0.241 \\
 &  & 2 & 0.947 & 39.9 & 0.897 & 14.4 & 0.973 & 0.338 \\
 &  & 4 & 0.944 & 40.3 & 0.912 & 19.2 & 0.962 & 0.447 \\
 &  & 8 & 0.938 & 40.4 & 0.914 & 20.8 & 0.956 & 0.481 \\
 &  & 12 & 0.933 & 40.5 & 0.912 & 21.2 & 0.950 & 0.487 \\
\midrule
ConvNeXt-T & C10N worst & 1 & 0.987 & 38.3 & 0.877 & 14.7 & 0.968 & 0.361 \\
 &  & 2 & 0.988 & 38.1 & 0.901 & 18.7 & 0.967 & 0.460 \\
 &  & 4 & 0.989 & 38.0 & 0.954 & 24.5 & 0.983 & 0.611 \\
 &  & 8 & 0.989 & 38.2 & 0.961 & 28.2 & 0.970 & 0.696 \\
 &  & 12 & 0.988 & 38.3 & 0.961 & 29.3 & 0.964 & 0.717 \\
\midrule
ConvNeXt-T & C100N & 1 & 0.918 & 40.5 & 0.821 & 8.7 & 0.937 & 0.196 \\
 &  & 2 & 0.920 & 39.8 & 0.833 & 9.0 & 0.937 & 0.204 \\
 &  & 4 & 0.910 & 39.5 & 0.874 & 13.6 & 0.944 & 0.310 \\
 &  & 8 & 0.907 & 39.6 & 0.890 & 16.1 & 0.944 & 0.367 \\
 &  & 12 & 0.904 & 39.7 & 0.892 & 16.7 & 0.941 & 0.380 \\
\bottomrule
\end{tabular}
\end{table}

\begin{table}[h]
\centering
\caption{Training precision and batch size (ViT-S, CIFAR-10N worst, true share 39.4\%, mean of five runs, 5{,}000 audited examples). Each row trains and audits with the stated precision and batch size, at the same learning rate and number of epochs. Fitted: share of the wrong labels that the final model predicts (\%). Reading in \%.}
\label{tab:abl-precbatch}
\small\setlength{\tabcolsep}{4pt}
\begin{tabular}{lccccccc}
\toprule
 & \multicolumn{2}{c}{Epoch 2} & \multicolumn{5}{c}{Epoch 40} \\
\cmidrule(lr){2-3}\cmidrule(lr){4-8}
Training & Reading & AUC & Fitted & Reading & AUC & Precision & Recall \\
\midrule
bfloat16, batch size 128 (as in the paper) & 38.8 & 0.992 & 98.6 & 28.9 & 0.965 & 0.973 & 0.716 \\
FP32, batch size 128 & 38.8 & 0.992 & 98.7 & 29.2 & 0.964 & 0.971 & 0.720 \\
bfloat16, batch size 64 & 38.5 & 0.992 & 99.6 & 29.0 & 0.965 & 0.975 & 0.719 \\
bfloat16, batch size 256 & 38.9 & 0.992 & 91.6 & 29.1 & 0.964 & 0.971 & 0.718 \\
\bottomrule
\end{tabular}
\end{table}

\begin{table}[h]
\centering
\caption{Sample size: AUC of the allocation margin of Eq.~\eqref{eq:margin} at the final model when the audit covers a random subset of the training set (mean of five runs and, below 50{,}000, of three random subsets each).}
\label{tab:abl-samples}
\small\setlength{\tabcolsep}{4pt}
\begin{tabular}{llccccc}
\toprule
 & & \multicolumn{5}{c}{Audited examples} \\
\cmidrule(lr){3-7}
Model & Labels & 500 & 1{,}000 & 2{,}000 & 5{,}000 & 50{,}000 \\
\midrule
ViT-S & C10N worst & 0.963 & 0.964 & 0.961 & 0.961 & 0.963 \\
ConvNeXt-T & C10N worst & 0.936 & 0.943 & 0.941 & 0.940 & 0.940 \\
ViT-S & C100N & 0.799 & 0.872 & 0.896 & 0.904 & 0.912 \\
ConvNeXt-T & C100N & 0.747 & 0.818 & 0.843 & 0.856 & 0.863 \\
\bottomrule
\end{tabular}
\end{table}

\begin{table}[h]
\centering
\caption{The label-swap audit from a random subset of the monitored coordinates, at the final model (5{,}000 audited
examples, on iNat the whole training set of 50{,}000, mean of five runs and, for the subsets, of three random subsets
each). iNat: 1{,}000 species of iNaturalist 2021 (Appendix~\ref{app:inat}). A subset is drawn once and used for
every example and class, and the allocation is normalised over the kept coordinates. Reading in \%.}
\label{tab:coords}
\small\setlength{\tabcolsep}{3.5pt}
\begin{tabular}{llrcccccc}
\toprule
 & & & \multicolumn{3}{c}{C10N worst} & \multicolumn{3}{c}{C100N} \\
\cmidrule(lr){4-6}\cmidrule(lr){7-9}
Model & Kept & Coordinates & AUC & Reading & Precision & AUC & Reading & Precision \\
\midrule
ViT-S & all & 2{,}952{,}192 & 0.965 & 28.9 & 0.973 & 0.912 & 19.2 & 0.962 \\
 & 10\% & 295{,}219 & 0.965 & 28.9 & 0.973 & 0.911 & 19.2 & 0.961 \\
 & 1\% & 29{,}522 & 0.960 & 29.1 & 0.965 & 0.905 & 19.6 & 0.954 \\
 & 0.1\% & 2{,}952 & 0.918 & 31.4 & 0.881 & 0.864 & 23.2 & 0.876 \\
\midrule
ResNet-50 & all & 3{,}407{,}872 & 0.958 & 30.8 & 0.948 & 0.905 & 19.8 & 0.942 \\
 & 10\% & 340{,}787 & 0.958 & 30.9 & 0.947 & 0.903 & 19.9 & 0.937 \\
 & 1\% & 34{,}079 & 0.952 & 31.7 & 0.936 & 0.888 & 21.0 & 0.908 \\
 & 0.1\% & 3{,}408 & 0.929 & 34.1 & 0.873 & 0.826 & 27.4 & 0.780 \\
\midrule
 & & & \multicolumn{3}{c}{iNat sym.\ 40\%} & \multicolumn{3}{c}{iNat asym.\ 40\%} \\
\cmidrule(lr){4-6}\cmidrule(lr){7-9}
ViT-S & all & 2{,}952{,}192 & 0.991 & 42.5 & 0.916 & 0.852 & 23.4 & 0.837 \\
 & 10\% & 295{,}219 & 0.991 & 42.5 & 0.916 & 0.851 & 23.4 & 0.836 \\
 & 1\% & 29{,}522 & 0.990 & 42.7 & 0.910 & 0.844 & 23.9 & 0.824 \\
\bottomrule
\end{tabular}
\end{table}

\begin{table}[h]
\centering
\caption{Coordinate subsets, per run: AUC of the label-swap score at the final model from all, 10\%, 1\% and 0.1\% of the monitored coordinates (5{,}000 audited examples, on iNat 50{,}000; for the subsets the mean of three random subsets).}
\label{tab:coords-perrun}
\small\setlength{\tabcolsep}{4pt}
\begin{tabular}{lllcccccc}
\toprule
Model & Labels & Kept & Run 1 & Run 2 & Run 3 & Run 4 & Run 5 & Mean \\
\midrule
ViT-S & C10N worst & all & 0.967 & 0.963 & 0.965 & 0.969 & 0.961 & 0.965 \\
 &  & 10\% & 0.967 & 0.962 & 0.965 & 0.968 & 0.960 & 0.965 \\
 &  & 1\% & 0.964 & 0.958 & 0.960 & 0.964 & 0.953 & 0.960 \\
 &  & 0.1\% & 0.933 & 0.911 & 0.930 & 0.926 & 0.892 & 0.918 \\
\midrule
ViT-S & C100N & all & 0.913 & 0.910 & 0.908 & 0.912 & 0.915 & 0.912 \\
 &  & 10\% & 0.912 & 0.909 & 0.907 & 0.911 & 0.913 & 0.911 \\
 &  & 1\% & 0.907 & 0.904 & 0.903 & 0.906 & 0.907 & 0.905 \\
 &  & 0.1\% & 0.865 & 0.868 & 0.852 & 0.867 & 0.866 & 0.864 \\
\midrule
ResNet-50 & C10N worst & all & 0.961 & 0.957 & 0.960 & 0.953 & 0.959 & 0.958 \\
 &  & 10\% & 0.960 & 0.956 & 0.960 & 0.952 & 0.959 & 0.958 \\
 &  & 1\% & 0.955 & 0.950 & 0.952 & 0.947 & 0.955 & 0.952 \\
 &  & 0.1\% & 0.933 & 0.929 & 0.929 & 0.928 & 0.928 & 0.929 \\
\midrule
ResNet-50 & C100N & all & 0.905 & 0.907 & 0.905 & 0.906 & 0.904 & 0.905 \\
 &  & 10\% & 0.903 & 0.905 & 0.903 & 0.904 & 0.901 & 0.903 \\
 &  & 1\% & 0.888 & 0.890 & 0.886 & 0.890 & 0.885 & 0.888 \\
 &  & 0.1\% & 0.826 & 0.829 & 0.822 & 0.828 & 0.825 & 0.826 \\
\midrule
ViT-S & iNat sym.\ 40\% & all & 0.991 & 0.991 & 0.991 & 0.991 & 0.991 & 0.991 \\
 &  & 10\% & 0.991 & 0.991 & 0.991 & 0.991 & 0.991 & 0.991 \\
 &  & 1\% & 0.990 & 0.990 & 0.990 & 0.990 & 0.990 & 0.990 \\
\midrule
ViT-S & iNat asym.\ 40\% & all & 0.853 & 0.852 & 0.850 & 0.852 & 0.852 & 0.852 \\
 &  & 10\% & 0.852 & 0.851 & 0.849 & 0.852 & 0.851 & 0.851 \\
 &  & 1\% & 0.845 & 0.844 & 0.841 & 0.845 & 0.843 & 0.844 \\
\bottomrule
\end{tabular}
\end{table}

\FloatBarrier
\section{Reproducibility of each experiment}
\label{app:repro}

\subsection{Main experiments}
\label{app:repro-main}

\begin{table}[!b]
\centering
\caption{Training configuration of every image setting. Wrong: share of wrong training labels. Learning rates in units of $10^{-5}$, selected by the search on the setting itself, or taken from CIFAR-10N worst (w) or from 50\% symmetric noise on CIFAR-100 (f) with the same backbone, or from symmetric noise on the same species (i). iNat: 1{,}000 species of iNaturalist 2021. All runs: AdamW, weight decay 0.05, batch size 128, linear warm-up over the first 10\% of steps, cosine decay, resolution 224. Asymmetric noise (asym.) moves 40\% of the labels of each affected class and is run with ViT-S.}
\label{tab:repro}
\small\setlength{\tabcolsep}{2.5pt}
\begin{tabular}{lrrcccc}
\toprule
Labels & Train images & Classes & Wrong (\%) & Epochs & ViT-S & ConvNeXt-T \\
\midrule
C10N aggregate & 50{,}000 & 10 & 9.0 & 40 & 2 & 5$^{\mathrm{w}}$ \\
C10N random-1 & 50{,}000 & 10 & 17.2 & 40 & 2 & 5$^{\mathrm{w}}$ \\
C10N worst & 50{,}000 & 10 & 40.2 & 40 & 2 & 5 \\
C10 sym.\ 20\% & 50{,}000 & 10 & 20.0 & 40 & 2$^{\mathrm{w}}$ & 5$^{\mathrm{w}}$ \\
C10 sym.\ 60\% & 50{,}000 & 10 & 60.0 & 40 & 2$^{\mathrm{w}}$ & 5$^{\mathrm{w}}$ \\
C10 asym.\ 40\% & 50{,}000 & 10 & 20.0 & 40 & 2$^{\mathrm{w}}$ & -- \\
C10 clean & 50{,}000 & 10 & 0 & 40 & 10 & 10 \\
C100N & 50{,}000 & 100 & 40.2 & 40 & 5 & 5 \\
C100 sym.\ 20\% & 50{,}000 & 100 & 20.0 & 40 & 7.5$^{\mathrm{f}}$ & 10$^{\mathrm{f}}$ \\
C100 sym.\ 60\% & 50{,}000 & 100 & 60.0 & 40 & 7.5$^{\mathrm{f}}$ & 10$^{\mathrm{f}}$ \\
C100 asym.\ 40\% & 50{,}000 & 100 & 40.0 & 40 & 7.5$^{\mathrm{f}}$ & -- \\
C100 clean & 50{,}000 & 100 & 0 & 40 & 10 & 15 \\
CIFAR-10H random-1 & 10{,}000 & 10 & 5.1 & 80 & 2$^{\mathrm{w}}$ & 5$^{\mathrm{w}}$ \\
CIFAR-10H clean & 10{,}000 & 10 & 0 & 80 & 2$^{\mathrm{w}}$ & 5$^{\mathrm{w}}$ \\
Food-101 sym.\ 40\% & 75{,}750 & 101 & 40.0 & 40 & 7.5$^{\mathrm{f}}$ & 10$^{\mathrm{f}}$ \\
Food-101N web & 52{,}867 & 101 & 18.4 & 40 & 7.5$^{\mathrm{f}}$ & 10$^{\mathrm{f}}$ \\
iNat sym.\ 40\% & 50{,}000 & 1{,}000 & 40.0 & 40 & 20 & -- \\
iNat asym.\ 40\% & 50{,}000 & 1{,}000 & 40.0 & 40 & 20$^{\mathrm{i}}$ & -- \\
\bottomrule
\end{tabular}
\end{table}

\paragraph{Data.} CIFAR-10N and CIFAR-100N \citep{wei2022cifarn} attach human annotations to the CIFAR training
images. We use the CIFAR-10N labels \emph{aggregate} (9.0\% wrong), \emph{random-1} (17.2\%) and \emph{worst}
(40.2\%), and the CIFAR-100N fine labels (40.2\%). Symmetric noise at 20\% and 60\% on CIFAR-10 and CIFAR-100, drawn with a fixed
seed, extends the range, asymmetric noise that moves labels to a similar class tests the case that annotators produce
most often (Appendix~\ref{app:asym}), and clean CIFAR-10 and CIFAR-100 serve as the no-noise condition. CIFAR-10H
\citep{peterson2019human} provides about 51 human annotations for each of the 10{,}000 CIFAR-10 test images. We train
on these images with the label of one annotator drawn at random for each image (5.1\% wrong), and with the clean
labels, and test on 10{,}000 images drawn from the CIFAR-10 training set. Food-101 \citep{bossard2014food} has 75{,}750
training and 25{,}250 test images of 101 dishes, and we add 40\% symmetric noise to its training labels. Food-101N
\citep{lee2018cleannet} collects web images for the same 101 classes, labelled by the search query, and a human
verification of the web label for part of them. We train on the 52{,}867 verified training images with their web labels
(18.4\% wrong according to the verification) and test on the Food-101 test set.

The learning rate is the one selected for cross-entropy on that data by a search on a noisy 10\% split of the training
set (for symmetric noise on CIFAR-10, for CIFAR-10H and, with ConvNeXt-T, for CIFAR-10N aggregate and random-1, the
rate selected for CIFAR-10N worst, and for symmetric noise on CIFAR-100 and for Food-101 and Food-101N, the rate
selected for 50\% symmetric noise on CIFAR-100). CIFAR-10H, with 10{,}000 training images, is trained for 80 epochs
and also kept after the last one.

\paragraph{Labels and audited rows.} Symmetric noise flips a class-stratified fraction of the training examples, drawn
once, each to a uniformly drawn other class. CIFAR-10N and CIFAR-100N labels are those of the official release,
matched to the CIFAR training index and checked against the official clean labels. For the random-annotator labels of
CIFAR-10H the annotator of each image is drawn once. The five runs of a setting use five consecutive training seeds and differ in
nothing else: the label noise and the 5{,}000 audited rows are the same in all of them and for every compared score.
Run $k$ uses the same seed in every setting.

\paragraph{Learning-rate search.} The learning rate is the only searched training coefficient. Candidates are
$\{1, 2, 5, 10, 20, 50\} \times 10^{-5}$. Each candidate is trained twice, with seeds disjoint from those of the
reported runs, for 4 epochs on 90\% of the training set and scored by its accuracy on the remaining 10\% with the
observed labels; the test set is never read. A candidate is excluded if a run fails or ends below twice the chance
accuracy. The best mean is then compared with the two rates halfway to its neighbours, and the best of the three is
kept. Table~\ref{tab:repro} lists the selected rates. Settings without a search of their own use the rate of the
setting marked in the table. ResNet-50 has a search of its own on CIFAR-10N worst and on CIFAR-100N, which selects
$2 \times 10^{-4}$ and $5 \times 10^{-4}$. For the highest candidate, the rate above it is taken at the same distance
as the rate halfway to the candidate below ($3.5$ and $6.5 \times 10^{-4}$). The 1{,}000 iNaturalist species have a
search of their own on symmetric noise. It excludes $10^{-5}$, which ends below twice the chance accuracy, and selects
$2 \times 10^{-4}$, which asymmetric noise on the same species also uses. The text models use the fixed rates of
Section~\ref{sec:text}.

\subsection{Extension experiments}
\label{app:repro-ext}

\paragraph{Text data and training.} NoisyAG-News \citep{noisyagnews} attaches three human labels to each of 50{,}000 AG News training
articles (four classes); we use its worst human labels (37.8\% wrong) and test on the AG News test set. For GLUE
\citep{wang2019glue} we add 40\% symmetric noise with a fixed seed to SST-2 and to fixed random subsets of 50{,}000
training pairs of QQP and MNLI, and test on the development sets (MNLI matched). We fine-tune RoBERTa-base
\citep{liu2019roberta} and DeBERTa-v3-base \citep{he2023debertav3} for 20 epochs (learning rate $2 \times 10^{-5}$,
batch size 32, weight decay 0.01, 6\% linear warm-up) and Qwen3-1.7B-Base \citep{qwen3} on NoisyAG-News for 6 epochs
(learning rate $10^{-5}$, batch size 16), each with a linear classification head, and audit the last four transformer
blocks of the final model on 5{,}000 training examples. Every setting is run five times.
Table~\ref{tab:text-repro} lists the configuration of every text setting and the test accuracy of its final models.

\begin{table}[h]
\centering
\caption{Configuration of every text setting. AG News, worst: the worst human labels of NoisyAG-News. Wrong: share of wrong training labels (\%). Test: size of the AG News test set or of the GLUE development set
(MNLI matched). Length: maximum number of tokens. Training: epochs, learning rate, batch size. Coordinates: monitored coordinates, the attention and MLP output projections of the last four blocks. All runs: AdamW, weight decay 0.01, linear warm-up over the first 6\% of steps, bfloat16, five runs. Accuracy: test accuracy of the final model (\%), mean (standard deviation) over the five runs.}
\label{tab:text-repro}
\footnotesize\setlength{\tabcolsep}{3pt}
\begin{tabular}{llrrcccccc}
\toprule
Model & Labels & Train & Test & Classes & Wrong & Length & Training & Coordinates & Accuracy \\
\midrule
RoBERTa & AG News, worst & 50{,}000 & 7{,}600 & 4 & 37.8 & 128 & 20, $2 \times 10^{-5}$, 32 & 11{,}802{,}624 & 68.1 (0.5) \\
 & SST-2, sym.\ 40\% & 67{,}349 & 872 & 2 & 40.0 & 64 & 20, $2 \times 10^{-5}$, 32 & 11{,}802{,}624 & 67.8 (0.3) \\
 & QQP, sym.\ 40\% & 50{,}000 & 40{,}430 & 2 & 40.0 & 128 & 20, $2 \times 10^{-5}$, 32 & 11{,}802{,}624 & 66.4 (8.3) \\
 & MNLI, sym.\ 40\% & 50{,}000 & 9{,}815 & 3 & 40.0 & 128 & 20, $2 \times 10^{-5}$, 32 & 11{,}802{,}624 & 63.0 (0.5) \\
\midrule
DeBERTa-v3 & AG News, worst & 50{,}000 & 7{,}600 & 4 & 37.8 & 128 & 20, $2 \times 10^{-5}$, 32 & 11{,}802{,}624 & 67.5 (0.8) \\
 & SST-2, sym.\ 40\% & 67{,}349 & 872 & 2 & 40.0 & 64 & 20, $2 \times 10^{-5}$, 32 & 11{,}802{,}624 & 67.9 (5.4) \\
 & QQP, sym.\ 40\% & 50{,}000 & 40{,}430 & 2 & 40.0 & 128 & 20, $2 \times 10^{-5}$, 32 & 11{,}802{,}624 & 60.1 (1.6) \\
 & MNLI, sym.\ 40\% & 50{,}000 & 9{,}815 & 3 & 40.0 & 128 & 20, $2 \times 10^{-5}$, 32 & 11{,}802{,}624 & 63.3 (0.7) \\
\midrule
Qwen3-1.7B & AG News, worst & 50{,}000 & 7{,}600 & 4 & 37.8 & 128 & 6, $10^{-5}$, 16 & 67{,}108{,}864 & 69.9 (0.4) \\
\bottomrule
\end{tabular}
\end{table}

\paragraph{Text models.} Inputs are truncated to 128 tokens (64 for SST-2), and the two sentences of a pair are encoded
together. The classification head reads the first token for RoBERTa and DeBERTa-v3 and the last non-padding token for
Qwen3-1.7B. All weights are fine-tuned.

\paragraph{Inversion attack.}
\label{app:privacy-setup}
The next two paragraphs give the inputs of the attack of Appendix~\ref{app:privacy}, and the attacker with its decoder.

\paragraph{Messages.} For each example the attack is given one of four inputs. (i) The message of the split audit:
the inner products of $q_i(y_i)$ and of $q_i(a_i)$ with the $K$ class sums, the two squared norms, and $y_i$ and $a_i$
(one-hot), computed exactly as in the audit. (ii) A sketch of the allocation $q_i(y_i)$ itself: each of its 2.95 million
coordinates is added with a random sign to one of 4{,}096 random buckets. (iii) The 384 final features that enter the
classifier head. (iv) The label alone: the reconstruction is the mean auxiliary image of the observed class, without a
decoder.

\paragraph{Attacker and decoder.} The attacker holds the trained model (the final model of each run), the owner's class
sums, which the owner never sends, and the test split of the dataset as labelled auxiliary images from the same
distribution. It computes the message of every auxiliary image with its label and trains, for each message, a decoder
from the standardised message to the image: two fully connected layers (1{,}024 units, then $256 \times 4 \times 4$) and
three stride-2 transposed convolutions to $3 \times 32 \times 32$ with a sigmoid output, mean squared error, Adam with
learning rate $10^{-3}$, batch size 256, 40 epochs. Images larger than $32 \times 32$ (Food-101 and Food-101N)
are area-averaged to $32 \times 32$ for training and scoring.

\paragraph{One thousand classes.} The classes are 1{,}000 of the 10{,}000 species of the iNaturalist 2021 competition
data \citep{vanhorn2021inat}, drawn once with a fixed seed. The training set holds the 50 images of each species in the
mini training split (50{,}000), and the test set its 10 validation images (10{,}000). The official labels are the
reference. The
1{,}000 species fall into 11 supercategories of 8 to 451 species each. Asymmetric noise moves 40\% of the labels of each
species to the next drawn species of its supercategory in label order, and those of the last one to the first.
Training follows Table~\ref{tab:repro}. Every audit covers the whole training set,
50 examples per class, with all 2{,}952{,}192 monitored coordinates or with a random 10\% or 1\% of them, three subsets
each. A run of 40 epochs takes about 20 minutes on one RTX 4090.

\subsection{Ablations}
\label{app:repro-abl}

Every ablation changes one thing and keeps the recipe of Table~\ref{tab:repro}, with five runs. The compared quantity and the
number $k$ of monitored blocks change only the audit, which is run again on the kept models. The precision and batch-size
rows train new models: full FP32 precision in training and in the audit, or batch size 64 or 256, at the learning rate and
number of epochs of the main recipe. In full precision the audit of 5{,}000 examples takes 28 instead of 14 seconds. With 1{,}000 classes one audit of all
50{,}000 examples takes 325 seconds with a peak GPU memory of 21.1 GiB, and 293 seconds and 11.2 GiB with 10\% of the
coordinates. The
sample-size rows audit random subsets of the training set, three subsets per size. A coordinate subset is drawn once per
subset seed and used for every example and class.

\FloatBarrier
\section{Protocol details common to all experiments}
\label{app:protocol}

\paragraph{Software and hardware.} All models are trained and audited with Python 3.12, PyTorch 2.5.1 and
\texttt{timm} 1.0.27; the text models use Hugging Face Transformers 5.8.0, and confident learning uses
\texttt{cleanlab} 2.7.1. The sign and Wilcoxon signed-rank tests use their exact two-sided null distributions.
The experiments run on NVIDIA RTX 4090 GPUs, with the forward and backward computations under bfloat16 autocast and
float32 weights (Appendix~\ref{app:impl}).\codestatement

\paragraph{Pretrained models.} The image models come from \texttt{timm} and the text models from the Hugging Face
Hub; a new linear classification head is attached to every model.
\begin{center}\small
\begin{tabular}{ll}
ViT-S/16 & \texttt{vit\_small\_patch16\_224.augreg\_in21k\_ft\_in1k} \\
ConvNeXt-T & \texttt{convnext\_tiny.fb\_in22k\_ft\_in1k} \\
ResNet-50 & \texttt{resnet50.a1\_in1k} \\
RoBERTa-base & \texttt{FacebookAI/roberta-base} \\
DeBERTa-v3-base & \texttt{microsoft/deberta-v3-base} \\
Qwen3-1.7B-Base & \texttt{Qwen/Qwen3-1.7B-Base} \\
\end{tabular}
\end{center}

\paragraph{Inputs.} CIFAR images are kept at $32 \times 32$. Food-101, Food-101N and iNaturalist images are resized to a
shorter side of 224 (bicubic) and centre-cropped to $224 \times 224$ once, before training. Training adds a random crop
with 4-pixel padding and a horizontal flip; every image is then resized bilinearly to the model resolution of 224 and
normalised with the mean and standard deviation of the pretrained model. The audit and all evaluations use the images
without augmentation.

\paragraph{Monitored parameters.} For ViT-S/16 the monitored layers are the attention output projection
(\texttt{attn.proj}, $384\times384$ weights plus bias) and the MLP output projection (\texttt{mlp.fc2},
$1536\times384$ plus bias) of the last four blocks, about 2.95 million coordinates in all. For ConvNeXt-T they are the
MLP output projections of the last four blocks, about 7.67 million coordinates, and for ResNet-50 the $1\times1$ output
convolutions (\texttt{conv3}, without bias) of the last four bottleneck blocks, about 3.41 million coordinates. For the
text models they are the attention output projection and the MLP output projection of the last four transformer blocks,
11{,}802{,}624 coordinates for RoBERTa-base and DeBERTa-v3-base (weights and biases) and 67{,}108{,}864 for Qwen3-1.7B
(weights only).
The classification head is never monitored.

\paragraph{Compared scores.} All scores are computed on the audited examples of the same frozen model. The loss is the
cross-entropy of the observed label. The feature margin is Eq.~\eqref{eq:margin} with the classifier's input features in
place of the allocations, the example left out of its own class mean. The neighbour vote is the share of an example's
ten nearest audited examples, by cosine similarity of these features, that carry another label, and it flags the
example when this share exceeds one half. The gradient norm is the $\ell_1$ norm of the gradient over the monitored
parameters and the self-influence is its squared $\ell_2$ norm, both under the observed label. TracIn is the sum of
the self-influence over the kept checkpoints, without learning-rate weights. Confident learning uses
\texttt{find\_label\_issues} of \texttt{cleanlab} with its default settings on the out-of-sample probabilities of the
whole training set, and its score is the predicted probability of the observed label.

\clearpage

\bibliographystyle{plainnat}
\bibliography{refs}
\end{document}